\documentclass[10pt,journal]{IEEEtran}

\usepackage[hyphens]{url}
\usepackage{graphicx}
\usepackage{amsmath}
\usepackage{amssymb}
\usepackage{xspace}
\usepackage{booktabs}
\usepackage{multirow}

\newcommand{\ours}{ExecDiffuser\xspace}
\newcommand{\igd}{IGD\xspace}
\newcommand{\egs}{EGS\xspace}

\title{ExecDiffuser: Closing the Sample-to-Execution Gap in Diffusion Planning}
\author{Shuo Pei$^{1,2}$, Bosun Liang$^{2}$, Saihui Zha$^{1}$, Chen Sun$^{2}$, Huachun Tan$^{1}$, and Yong Wang$^{2,*}$%
\thanks{$^{1}$Shuo Pei, Saihui Zha, Huachun Tan are with the Beijing Institute of Technology, Guangdong, China.
$^{2}$Shuo Pei, Bosun Liang, Chen Sun, and Yong Wang (Corresponding author) are with the Department of Data and Systems Engineering, The University of Hong Kong, Hong Kong, China. (e-mail: wy0304@hku.hk).}}

\begin{document}
\maketitle

\begin{abstract}
Generative planners are often evaluated by the trajectories they can generate, although closed-loop driving ultimately depends on the single candidate committed to control. We identify this mismatch as the \emph{sample-to-execution gap} and show that improving proposal quality alone
can worsen the executed decision. Here, we introduce \ours, a recurrent proposal-to-execution framework that couples candidate generation, trajectory selection, and subsequent replanning through shared execution semantics. Intent-Guided Diffusion Proposal (\igd) assigns hypothesis responsibility and shapes trajectories with execution-conditioned supervision. Execution-Grounded Selection (\egs) learns a candidate-conditioned publication utility, commits one trajectory to control, and returns it as context for the next planning cycle. Across nuPlan, InterPlan, and NAVSIM, \ours achieves leading closed-loop performance against strong state-of-the-art baselines. Under a unified
five-seed protocol, it obtains an average score of 85.62, improving over Diffusion Planner and FlowDrive by 4.02 and 1.31 points, respectively. On NAVSIM, it achieves 92.6 PDMS under the official evaluation pipeline. Matched counterfactual analysis further shows that \ours more
reliably converts strong candidates into committed decisions, while prefix interventions confirm that this advantage persists after physical execution
and subsequent replanning.
\end{abstract}

\section{Introduction}
\label{sec:introduction}

Autonomous driving planners repeatedly use sensory observations, map
and route context, and recent vehicle motion to produce a future ego
trajectory for downstream control~\cite{pei2026safe,wang2026q,
tang2024using,hu2023uniad}. In closed-loop operation, however, the
environment responds only to the trajectory committed at the current
planning cycle. This commitment determines the vehicle's immediate
motion and changes the interaction state from which the planner must
subsequently replan.

Generative planners are attractive because uncertain and interactive
scenes often admit multiple plausible maneuvers. Diffusion- and
flow-based planners therefore produce candidate sets rather than a
single deterministic trajectory. Recent work improves proposal quality
through flexible guidance, multimodal anchors, latent or bridge
representations, reinforcement fine-tuning, and flow matching
~\cite{zheng2025diffusionplanner,liao2025diffusiondrive,
liu2026bridgedrive,zhang2025lap,li2026plannerrft,
zou2025diffusiondrivev2,tan2025flowplanner,wang2025flowdrive}.
Recent variants also reduce sampling redundancy, introduce
spatiotemporal or scene-adaptive guidance, and explicitly improve
trajectory feasibility~\cite{li2025flashplanner,yu2026g2dp,
wu2026cddrive,wang2026feaxdrive}. These advances raise the quality and
coverage of the candidate set, but only one trajectory is ultimately
committed to control. A better proposal set therefore need not produce
a better executed decision.

This distinction exposes a gap between proposal generation and
trajectory commitment. A planner may already contain a strong maneuver
in its candidate set, yet commit a weaker alternative because the
selector was not trained to compare realized candidates under the
current execution context. We refer to this mismatch as the
\emph{sample-to-execution gap}. A controlled diagnosis reveals a
counterintuitive failure. With the imitation-based selector fixed,
stronger proposal learning raises counterfactual oracle-best quality
from 90.33 to 91.36, but lowers committed quality from 89.82 to 89.64.
As a result, candidate-publication regret increases from 0.51 to
1.72. Thus, improving the proposal ceiling alone can make the final
closed-loop decision worse.

The underlying cause is a mismatch between how proposals are organized
and how decisions are executed. Proposal learning typically assigns
responsibility to latent hypotheses according to imitation proximity
or geometric matching. Commitment, in contrast, must rank the realized
trajectories using evidence available at the current planning cycle,
including driveability, route adherence, predicted interactions,
progress, comfort, and recent execution history. When these stages are
optimized independently, the hypotheses that receive proposal
supervision are not necessarily aligned with the evidence required for
controller-facing selection. The problem is inherently recurrent: each
committed trajectory changes the vehicle and interaction state, which
in turn affects subsequent planning. Candidate generation, trajectory
commitment, and replanning should therefore be connected through a
common execution interface.

\begin{figure}[!t]
    \centering
    \includegraphics[width=\columnwidth]{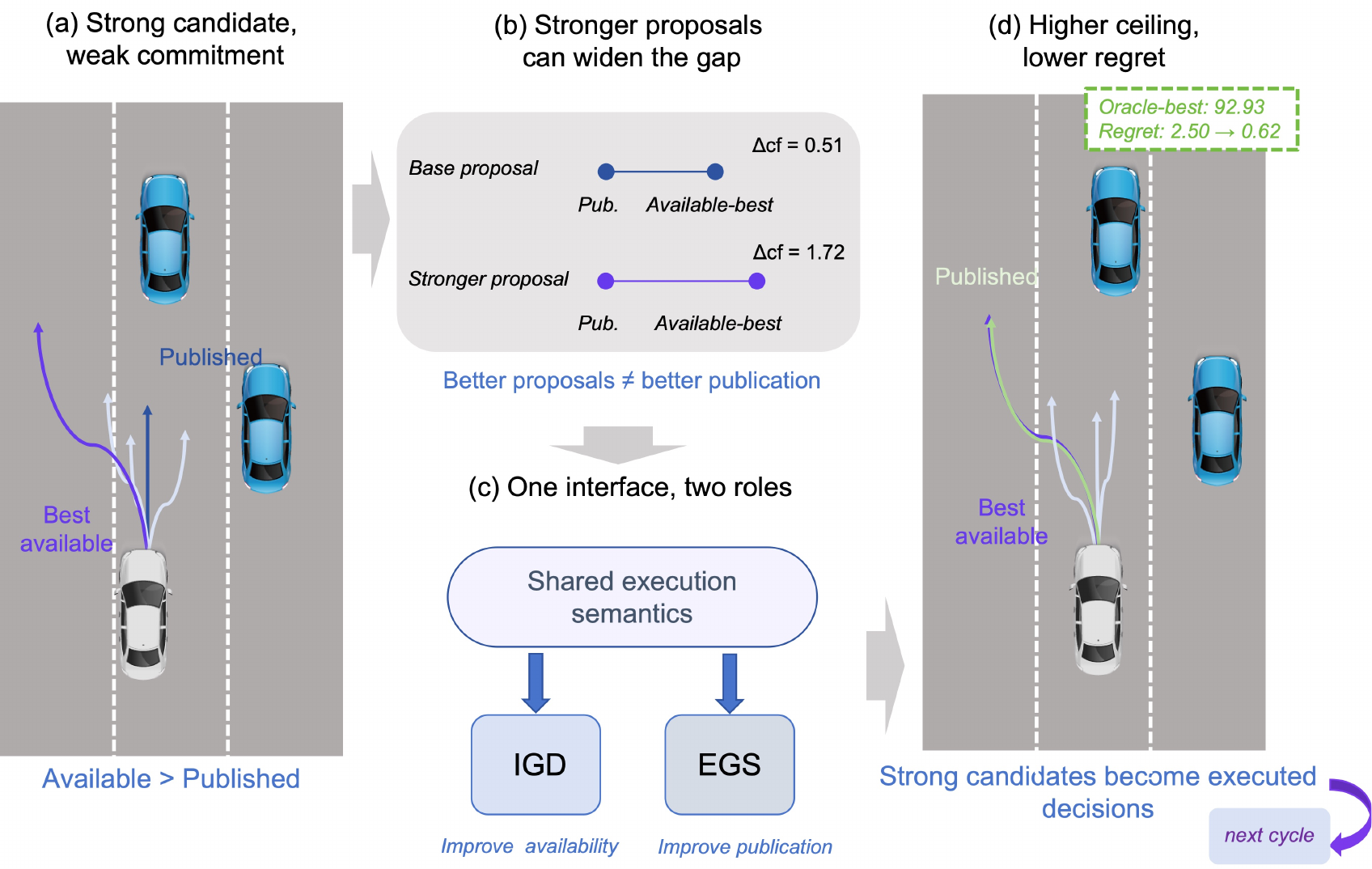}
    \caption{
    A strong candidate can already be available yet remain uncommitted
    to control. Strengthening proposals alone may raise the
    available-candidate ceiling while widening candidate-publication
    regret. \ours connects proposal learning and trajectory publication
    through shared execution semantics: \igd improves candidate
    availability, while recurrent \egs commits one trajectory to
    control and returns it to the next planning cycle.
    }
    \label{fig:overview}
\end{figure}

We introduce \ours, a recurrent framework that connects proposal
learning and controller-facing publication through shared execution
semantics. \igd improves candidate availability through
execution-conditioned responsibility and trajectory shaping, while
\egs learns a candidate-conditioned publication utility and returns
the selected trajectory as recurrent context. We separate these effects
through matched counterfactual evaluation and prefix interventions that
test whether the initial publication alters subsequent closed-loop
replanning. Under a unified five-seed protocol, \ours outperforms the
reproduced Diffusion Planner and FlowDrive baselines in all six nuPlan
settings, with consistent gains on interaction-intensive InterPlan and strong
performance on NAVSIM. Our contributions are
threefold:

\begin{itemize}
    \item We formulate the \emph{sample-to-execution gap} and introduce
    a recurrent proposal-to-execution architecture that couples
    candidate generation, controller-facing publication, and
    next-cycle replanning in closed loop.

    \item We develop shared execution semantics with distinct roles on
    both sides of the gap. \igd aligns hypothesis responsibility and
    trajectory shaping with execution-conditioned supervision, while
    recurrent \egs learns which realized candidate to commit under the
    current scene and execution history.

    \item We establish a matched counterfactual evaluation protocol
    that separates available-best quality from published quality.
    Commit-horizon controls and resumed-replanning interventions further
    establish that publication differences persist after physical
    execution and alter subsequent closed-loop planning.
\end{itemize}

\section{Related Work}
\label{sec:related_work}

\paragraph{Generative trajectory planning.}
Generative planners model driving as multimodal trajectory generation.
Diffusion Planner formulates planning as conditional trajectory
denoising~\cite{zheng2025diffusionplanner}, while DiffusionDrive
combines multimodal anchors with truncated diffusion for efficient
sampling~\cite{liao2025diffusiondrive}. BridgeDrive improves proposal
consistency through a diffusion bridge~\cite{liu2026bridgedrive}, and
LAP performs generation in a compact latent space
~\cite{zhang2025lap}. PlannerRFT and DiffusionDriveV2 further introduce
reward- or closed-loop-oriented supervision
~\cite{li2026plannerrft,zou2025diffusiondrivev2}, whereas FlowPlanner
and FlowDrive use flow-based dynamics for efficient multimodal
generation~\cite{tan2025flowplanner,wang2025flowdrive}. These methods
substantially improve the quality, diversity, and efficiency of
trajectory proposals. Yet a final closed-loop score does not reveal the
available-candidate ceiling before one trajectory is committed to
control.

\paragraph{Trajectory selection and closed-loop publication.}
Closed-loop planners increasingly evaluate trajectories using
structured scene context, interaction reasoning, and planning-oriented
objectives rather than open-loop imitation error alone. PlanTF analyzes
architectural and training choices that shape closed-loop imitation
planning~\cite{cheng2024plantf}; PLUTO combines structured scene
representations with auxiliary planning objectives
~\cite{cheng2024pluto}; and GameFormer jointly models interactive
prediction and planning~\cite{huang2023gameformer}. PDM-Closed directly
demonstrates the disagreement between open-loop accuracy and closed-loop
driving quality and selects among trajectory proposals using
planning-oriented objectives~\cite{dauner2023misconceptions}. These
works establish the importance of execution-aware trajectory
evaluation, but they do not isolate whether closed-loop performance is
limited by candidate availability or by the decision that commits one
candidate to control.

The proposal--publication coupling is already shaped during multimodal
supervision. MultiPath and CoverNet represent multiple futures through
trajectory anchors or discrete trajectory sets
~\cite{chai2020multipath,phanminh2020covernet}, while winner-takes-all
objectives encourage hypothesis specialization~\cite{xu2025awta}.
Assignment determines which latent hypothesis receives proposal
supervision, whereas candidate scoring determines which realized
trajectory is executed. Separate assignment and scoring objectives do
not explicitly align proposal responsibility with the evidence used
for final execution. \ours connects the two stages through shared
execution semantics: \igd aligns proposal responsibility and trajectory
shaping with execution evidence, while recurrent \egs uses the same
semantics to learn which generated candidate to commit. A matched
counterfactual evaluator separately measures candidate availability and
published quality.

\section{Method}
\label{sec:method}

\subsection{Problem Formulation and Overview}
\label{sec:method_overview}

At cycle $t$, the planner receives scene, map, route, and
previous-publication context
$(\mathcal O_t,\mathcal M_t,\mathcal R_t,
\tau_{t-1}^{\mathrm{pub}})$. The proposal model generates candidates,
from which \egs publishes one trajectory:
\begin{equation}
\begin{aligned}
\mathcal T_t
&=G_\theta(\mathcal O_t,\mathcal M_t,\mathcal R_t,
\tau_{t-1}^{\mathrm{pub}},K_t)
=\{\tau_t^{(k)}\}_{k=1}^{K_t},\\
k_t^\ast
&=V_\phi(\mathcal T_t,\mathcal O_t,\mathcal M_t,\mathcal R_t,
\tau_{t-1}^{\mathrm{pub}}),\qquad
\tau_t^{\mathrm{pub}}=\tau_t^{(k_t^\ast)}.
\end{aligned}
\label{eq:planning_interface}
\end{equation}

Let $\mathcal V_t=\{k:m_t^{(k)}=1\}$ denote the candidates admitted
by the online mask, and let
$\Omega_{\mathrm{sel}}=\{t:\mathcal V_t\neq\varnothing\}$ denote
the states in which EGS publishes a learned candidate. For retrospective
utility $J_t$, evaluated from the same initial state for every candidate,
the sample-to-execution gap is
\begin{equation}
\mathcal G_{\mathrm{S2E}}
=
\mathbb E_{t\in\Omega_{\mathrm{sel}}}
\left[
\max_k J_t(\tau_t^{(k)})
-J_t(\tau_t^{\mathrm{pub}})
\right].
\label{eq:sample_execution_gap}
\end{equation}
Equation~\eqref{eq:sample_execution_gap} isolates the publication
side of the sample-to-execution gap. The counterfactual oracle measures
the quality already available in the complete proposal set, whereas
$J_t(\tau_t^{\mathrm{pub}})$ measures the trajectory committed to
control. We call their difference \emph{candidate-publication regret}.
Because the oracle includes candidates rejected by the online mask, the
diagnostic captures both false rejection and ranking errors; all-invalid
states are reported separately. Since $J_t$ depends on future execution,
it is used only retrospectively. Prefix execution and resumed replanning
test whether the measured gap affects subsequent closed-loop behavior.

\begin{figure*}[!t]
    \centering
    \includegraphics[width=0.99\textwidth]{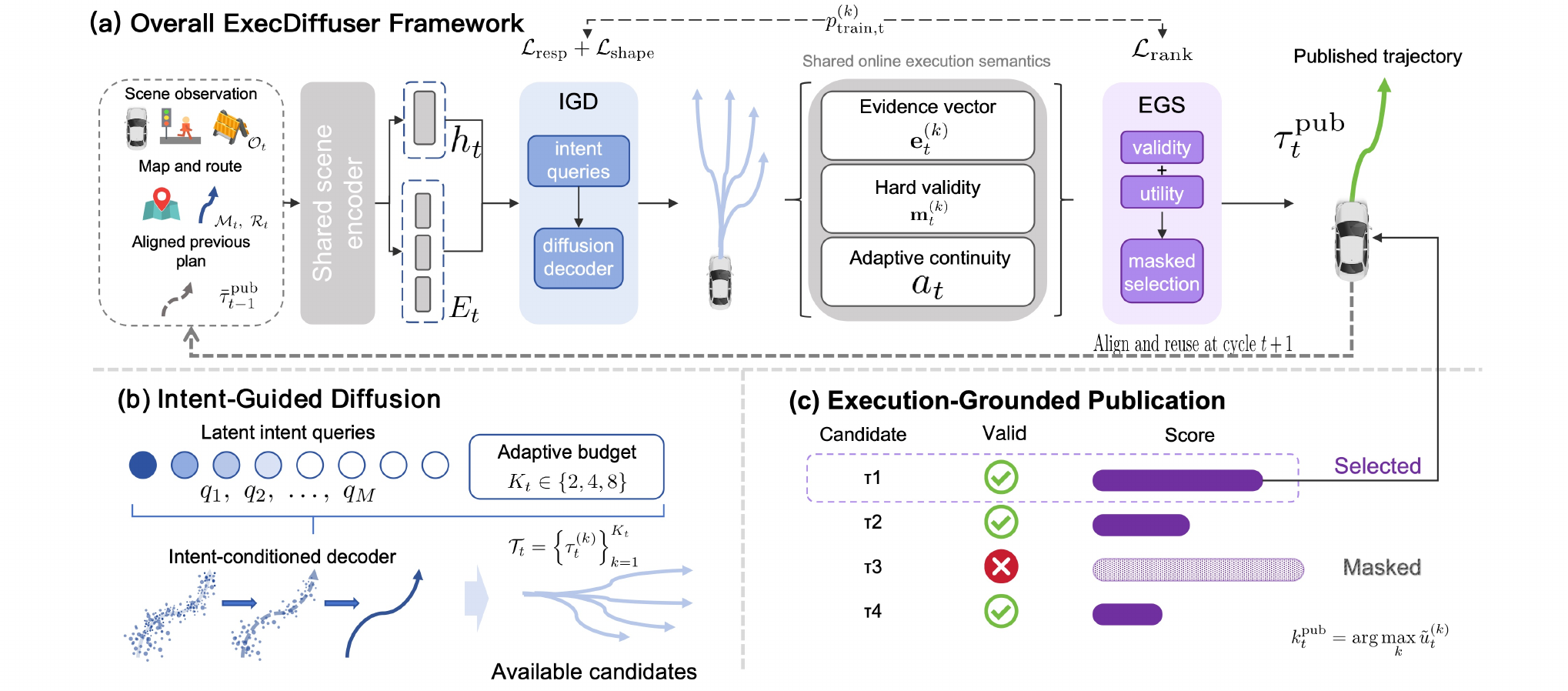}
    \caption{\ours couples proposal learning and publication through
    a shared online execution interface. \igd assigns and shapes latent
    hypotheses; \egs filters invalid candidates, ranks the remaining
    candidates, publishes one trajectory, and returns it as recurrent
    context.}
    \label{fig:framework}
\end{figure*}

Figure~\ref{fig:framework} summarizes the shared interface: \igd
separates hypothesis ownership from trajectory correction, improving
candidate availability, whereas \egs ranks realized candidates and
returns the publication as next-cycle context. A shared encoder yields
$h_t$ and $E_t$. Route context enters \igd through $c_t$ and \egs
through $e_{\mathrm{route},t}^{(k)}$, while
$\hat{\mathcal A}_{t,1:T}$ supplies risk evidence. The unexecuted suffix
of the previous publication is ego-aligned as
$\bar{\tau}_{t-1}^{\mathrm{pub}}$.

\subsection{Shared Online Execution Interface}
\label{sec:execution_interface}

The interface uses only information available at the current planning
cycle. A lightweight head predicts nearby-agent futures
$\hat{\mathcal A}_{t,1:T}$ from $(h_t,E_t)$. For candidate
$\tau_t^{(k)}$, it returns
\begin{equation}
\mathcal I_t\!\left(\tau_t^{(k)}\right)
=\left(\mathbf e_t^{(k)},m_t^{(k)},a_t\right),
\qquad
\mathbf e_t^{(k)}=[e_{t,d}^{(k)}]_{d\in\mathcal D_e},
\label{eq:execution_interface}
\end{equation}
where $\mathbf e_t^{(k)}$ contains normalized drive, route, lane, risk,
progress, kinematic, comfort, and temporal evidence; smaller values
are preferable. The hard mask is deliberately narrow:
\begin{equation}
\begin{aligned}
m_t^{(k)}
&=
\mathbb I\!\left[
\mathrm{Drivable}\!\left(\tau_t^{(k)}\right)
\right] \\
&\quad\cdot
\mathbb I\!\left[
\neg\mathrm{Collision}_{\mathrm{imm}}
\!\left(
\tau_t^{(k)},\hat{\mathcal A}_{t,1:T}
\right)
\right].
\end{aligned}
\label{eq:validity}
\end{equation}
$\mathrm{Drivable}$ checks the swept ego footprint and ego-center
tolerance. $\mathrm{Collision}_{\mathrm{imm}}$ compares
uncertainty-inflated predicted agent footprints with calibrated overlap
and TTC thresholds over a fixed look-ahead horizon. Route progress,
comfort, graded risk, and temporal consistency remain soft preferences.
A continuity gate $a_t\in[0,1]$ activates temporal evidence only while
the aligned previous publication remains suitable. For the analytic
rule-cost control, the same online evidence defines
\begin{equation}
\mathcal C_{\mathrm{rule},t}^{(k)}
=
\lambda_{\mathrm{exec}}\mathbf w^\top
\mathbf e_{t,\mathrm{nt}}^{(k)}
+
\lambda_{\mathrm{temp}}a_t e_{\mathrm{temp},t}^{(k)},
\label{eq:online_rule_cost}
\end{equation}
where $\mathbf e_{t,\mathrm{nt}}^{(k)}$ excludes temporal evidence.
The rule-cost baseline minimizes Eq.~\eqref{eq:online_rule_cost} over
admitted candidates. Full EGS instead learns publication utility from
candidate, scene, evidence, and recurrent context. The shared evidence
assigns proposal responsibility, supplies differentiable shaping terms,
and conditions EGS. No future simulator state or candidate-specific
reactive rollout is available online. The Appendix gives
the complete definitions and calibration values.

\subsection{Intent-Guided Diffusion Proposal}
\label{sec:igd}

\igd maintains $M$ learnable intent queries
$\mathcal Q=\{q_m\}_{m=1}^{M}$. Each active query receives route and
recurrent context, is paired with one noise sample, and is decoded into
a physical trajectory:
\begin{equation}
\begin{aligned}
c_t
&=f_{\mathrm{prior}}\!\left(
E_t,\mathcal R_t,\bar{\tau}_{t-1}^{\mathrm{pub}}\right),\\
g_t^{(m)}
&=f_{\mathrm{fuse}}\!\left(h_t,q_m,c_t\right),\\
\tau_t^{(m)}
&=\Pi\!\left(D_\theta\!\left(
\mathbf x_{\ell,t}^{(m)},\ell,g_t^{(m)},E_t\right)\right).
\end{aligned}
\label{eq:intent_diffusion}
\end{equation}
Here $D_\theta$ predicts a normalized clean-state sequence, and $\Pi$ is
a fixed differentiable reconstruction operator. \igd separates
hypothesis ownership from trajectory correction. It derives a training
preference over query--noise hypotheses from
\begin{equation}
\begin{aligned}
\mathcal A_{\mathrm{train},t}^{(k)}
={}&\mathcal L_{\mathrm{imit}}\!\left(
\tau_t^{(k)},\tau_t^{\mathrm{gt}}\right)\\
&+\lambda_{\mathrm{exec}}
\mathbf w^\top\mathbf e_{t,\mathrm{nt}}^{(k)}
+\lambda_{\mathrm{temp}}a_t e_{\mathrm{temp},t}^{(k)}\\
&+\lambda_{\mathrm{invalid}}\!\left(1-m_t^{(k)}\right),\\
p_{\mathrm{train},t}^{(k)}
={}&
\frac{\exp(-\mathcal A_{\mathrm{train},t}^{(k)}/\eta)}
{\sum_j\exp(-\mathcal A_{\mathrm{train},t}^{(j)}/\eta)}.
\end{aligned}
\label{eq:execution_responsibility}
\end{equation}
Here, $\mathbf e_{t,\mathrm{nt}}^{(k)}$ contains the normalized online
evidence of Eq.~\eqref{eq:online_rule_cost}, excluding its temporal
dimension, and $\eta$ is annealed during warm-up. Unlike the online rule
cost, $\mathcal A_{\mathrm{train},t}^{(k)}$ includes ground-truth
imitation proximity and is available only during training. The detached
$p_{\mathrm{train},t}^{(k)}$ assigns hypothesis ownership; a separate
differentiable path corrects trajectory geometry:
\begin{equation}
\begin{aligned}
\mathcal L_{\mathrm{resp}}
&=\sum_k\operatorname{sg}[p_{\mathrm{train},t}^{(k)}]
\mathcal L_{\mathrm{diff}}^{(k)},\\
\mathcal L_{\mathrm{shape}}
&=\sum_k\operatorname{sg}[p_{\mathrm{train},t}^{(k)}]
\widetilde{\mathcal C}_{\mathrm{shape}}(\tau_t^{(k)}).
\end{aligned}
\label{eq:proposal_losses}
\end{equation}
$\mathcal L_{\mathrm{resp}}$ assigns the query--noise hypothesis,
whereas $\mathcal L_{\mathrm{shape}}$ updates its physical trajectory.
The objective $\widetilde{\mathcal C}_{\mathrm{shape}}$ contains
coordinate-differentiable surrogates of route, lane, soft-overlap,
progress, kinematic, comfort, and temporal evidence; discrete collision
and off-road indicators remain detached. Specialization regularizes
query usage, and a scene-conditioned allocation head predicts
$K_t\in\{2,4,8\}$ before decoding the highest-priority queries. The
The Appendix gives the surrogate objective, auxiliary losses,
and calibration.

\subsection{Recurrent Execution-Grounded Selection}
\label{sec:egs}

For each candidate, \egs encodes the detached trajectory, attends to
the scene, and combines the result with the previous publication,
execution evidence, and continuity features:
\begin{equation}
\begin{aligned}
\mathbf z_t^{(k)}
=\operatorname{Concat}\big[
&f_\tau(\operatorname{sg}[\tau_t^{(k)}]),
f_{\mathrm{attn}}(\operatorname{sg}[\tau_t^{(k)}],E_t),\\
&f_{\mathrm{prev}}(\bar{\tau}_{t-1}^{\mathrm{pub}}),
f_e(\mathbf e_t^{(k)}),a_t\big],\\
u_\phi^{(k)}
={}&u_\phi(\mathbf z_t^{(k)}).
\end{aligned}
\label{eq:candidate_score}
\end{equation}
The scorer is trained against the same detached training preference:
\begin{equation}
\mathcal L_{\mathrm{rank}}
=-\sum_k\operatorname{sg}\!\left[p_{\mathrm{train},t}^{(k)}\right]
\log
\frac{\exp(u_\phi^{(k)}/\xi)}
{\sum_j\exp(u_\phi^{(j)}/\xi)}.
\label{eq:ranking_loss}
\end{equation}
The detached training preference supervises relative publication
quality, but it is not an inference-time energy and sends no coordinate
gradient through the ranking path. \egs instead infers a
context-dependent utility from online-observable candidate geometry,
scene tokens, route and forecast evidence, and the previous publication.
It does not reproduce $\mathcal C_{\mathrm{rule},t}^{(k)}$, which is
restricted to a fixed linear aggregation of predefined evidence. Among
admitted candidates, \egs publishes
\begin{equation}
k_t^\ast
=\arg\max_{k\in\mathcal V_t}u_\phi^{(k)},
\qquad
\tau_t^{\mathrm{pub}}=\tau_t^{(k_t^\ast)}.
\label{eq:inference_selection}
\end{equation}
If $\mathcal V_t=\varnothing$, the planner invokes a deterministic
jerk-limited MRM and records an all-invalid state. The selected
trajectory is stored, aligned, and returned to \igd and \egs at the
next cycle; gradients do not pass across planning cycles. The full
objective adds specialization, allocation, continuity-gate, and
forecast losses to Eqs.~\eqref{eq:proposal_losses}
and~\eqref{eq:ranking_loss}; their definitions are provided in the
Appendix.

\section{Experiments}
\label{sec:experiments}
We organize the experiments around four questions:
(1) Does ExecDiffuser improve closed-loop planning under a unified reproduction protocol?
(2) Does IGD improve the quality and validity of the available candidate set, independently of candidate selection?
(3) Does EGS improve the conversion of available candidates into committed decisions, independently of proposal generation?
(4) Does the initial trajectory commitment alter the subsequent closed-loop state and replanning outcome?

\subsection{Experimental Setup}
\label{sec:experimental_setup}

We evaluate nuPlan~\cite{karnchanachari2024nuplan} Val14, Test14-hard,
and Test14 under non-reactive (NR) and reactive (R) closed-loop
simulation, together with InterPlan~\cite{hallgarten2024interplan} for
interaction-intensive long-tail behavior. We reproduce Diffusion
Planner and FlowDrive as the closest diffusion- and flow-based
generative baselines. Diffusion Planner, FlowDrive, and \ours use the
same nuPlan version, scenario lists, simulator settings, metric
implementation, and five training seeds, without post-hoc trajectory
refinement. Results for other methods are quoted from their original
papers and used only as context.

For candidate-level diagnosis, we clone each eligible simulator state
and execute every generated candidate independently for the same
8.0-s horizon without replanning. NR branches share recorded future
agent states; R branches reset the simulator and reactive-agent state
before every rollout. Selected and oracle-best candidates therefore
share the initial simulator state, random seed, horizon, reactive-agent
model, and metric aggregation. Candidate-publication regret follows
Eq.~\eqref{eq:sample_execution_gap} over $\Omega_{\mathrm{sel}}$.
The oracle spans the complete generated set, including candidates
rejected online, while all-invalid/MRM states are reported separately.
Oracle top-1 recall counts a hit whenever the published trajectory is
among the counterfactual maximizers.

\ours observes 2.0~s of history, plans 8.0~s ahead at 10~Hz, uses eight
intent queries, selects $K\in\{2,4,8\}$ adaptively, and runs 25
denoising steps at inference. All reproduced methods are trained with
seeds $\{0,1,2,3,4\}$, and we report the arithmetic mean without seed
selection or ensembling. Confidence intervals use 10,000 paired
scenario-level bootstrap resamples with matched scenario seeds;
planning states are never treated as independent samples.
Appendices~A and~B
provide the complete configuration, diagnostic-state construction,
per-seed results, and statistical protocol.

\subsection{Closed-Loop Results on nuPlan and InterPlan}
\label{sec:main_results}

\begin{table*}[!t]
\centering
\small
\setlength{\tabcolsep}{3.2pt}
\begin{tabular}{lcccccccc}
\toprule
Method &
Val14 NR &
Val14 R &
Hard NR &
Hard R &
Test14 NR &
Test14 R &
Avg. &
InterPlan \\
\midrule
\multicolumn{9}{l}{
\textit{Results reported in the original papers for context}} \\
PDM-Closed~\cite{dauner2023misconceptions}
& 92.84 & 92.12 & 65.08 & 75.19 & 90.05 & 91.63 & 84.49 & -- \\
PlannerRFT~\cite{li2026plannerrft}
& 89.96 & 84.46 & 77.16 & 72.21 & -- & -- & -- & -- \\
LAP~\cite{zhang2025lap}
& 89.37 & 82.23 & 78.52 & 70.53 & 90.64 & 85.32 & 82.77 & -- \\
G2DP~\cite{yu2026g2dp}
& 92.26 & 83.91 & 80.05 & 73.91 & 91.58 & 85.26 & 84.50 & -- \\
G2DP, no history~\cite{yu2026g2dp}
& 91.08 & 86.67 & 78.77 & 77.61 & 88.54 & 88.42 & 85.18 & -- \\
\midrule
\multicolumn{9}{l}{
\textit{Five-seed means under our unified reproduction protocol}} \\
Diffusion Planner~\cite{zheng2025diffusionplanner}
& 89.87 & 82.38 & 75.99 & 69.22 & 89.19 & 82.93 & 81.60 & 24.71 \\
FlowDrive~\cite{wang2025flowdrive}
& 91.21 & 85.37 & 77.86 & 73.09 & 91.02 & 87.28 & 84.31 & 36.96 \\
\textbf{\ours}
& \textbf{92.37}
& \textbf{87.11}
& \textbf{79.13}
& \textbf{74.61}
& \textbf{92.16}
& \textbf{88.32}
& \textbf{85.62}
& \textbf{38.35} \\
\bottomrule
\end{tabular}
\normalsize
\caption{
Closed-loop results on nuPlan and InterPlan. NR/R denote
non-reactive/reactive simulation. Bold compares methods reproduced
under the same five-seed protocol; the remaining rows provide
reported-paper context.
}
\label{tab:main_results}
\end{table*}

Table~\ref{tab:main_results} answers the first question. ExecDiffuser
achieves the best overall performance across the six nuPlan settings,
with an average score of 85.62. Under the unified five-seed
reproduction protocol, it improves over Diffusion Planner and FlowDrive
by 4.02 and 1.31 points, respectively. The gains are largest in
reactive simulation: ExecDiffuser exceeds Diffusion Planner by 4.73
points on Val14-R and 5.39 points on Test14-hard-R, and FlowDrive by
1.74 and 1.52 points, respectively. This stronger improvement in
reactive settings is particularly relevant because they capture how
the committed ego trajectory changes subsequent interactions, whereas
non-reactive simulation primarily evaluates its immediate trajectory
quality.

On interaction-intensive InterPlan, \ours reaches 38.35, improving Diffusion Planner by 55.2\% and FlowDrive by 3.8\%. Together with the consistent nuPlan gains, this result positions the contribution as improved proposal-to-execution conversion rather than a benchmark gain
confined to one simulator mode. Complete per-seed results and paired comparisons are reported in Appendix Sec.~B.

\subsection{Ablation and Closed-Loop Diagnosis}
\label{sec:mechanism_ablation}

\begin{table*}[!t]
\centering
\small

\begin{minipage}[t]{0.485\textwidth}
\centering
\textbf{(a) Proposal learning with EGS fixed}\\[3pt]

\resizebox{\linewidth}{!}{%
\begin{tabular}{@{}lcccc@{}}
\toprule
Proposal training &
\shortstack{Val14\\NR/R $\uparrow$} &
\shortstack{Oracle-best\\$\uparrow$} &
\shortstack{Feasible ratio\\(\%) $\uparrow$} &
\shortstack{All-invalid\\(\%) $\downarrow$} \\
\midrule
IGD + imitation WTA
& 90.84/83.95 & 91.36 & 86.60 & 2.17 \\
+ resp. + shaping
& 92.05/86.41 & 92.63 & 92.10 & 0.82 \\
+ specialization
& \textbf{92.37/87.11}
& \textbf{92.93}
& \textbf{93.40}
& \textbf{0.68} \\
\bottomrule
\end{tabular}%
}
\end{minipage}
\hfill
\begin{minipage}[t]{0.485\textwidth}
\centering
\textbf{(b) Publication with proposals fixed}\\[3pt]

\resizebox{\linewidth}{!}{%
\begin{tabular}{@{}lcccc@{}}
\toprule
Selector &
\shortstack{Val14\\NR/R $\uparrow$} &
\shortstack{TTC score\\$\uparrow$} &
\shortstack{Top-1 recall\\(\%) $\uparrow$} &
\shortstack{Regret\\$\downarrow$} \\
\midrule
Imitation selector
& 90.49/83.29 & 88.24 & 56.4 & 2.50 \\
Rule-cost selector
& 91.17/84.18 & 90.12 & 68.9 & 1.82 \\
+ score/evidence/validity
& 92.15/86.14 & 92.89 & 82.4 & 0.84 \\
+ recurrence (EGS)
& \textbf{92.37/87.11}
& \textbf{94.06}
& \textbf{86.6}
& \textbf{0.62} \\
\bottomrule
\end{tabular}%
}
\end{minipage}

\normalsize
\caption{
Stage-controlled Val14 ablations. Panel (a) fixes the full EGS
selector and changes proposal learning; panel (b) freezes the same full
proposal generator and changes publication. Rule cost minimizes
Eq.~\eqref{eq:online_rule_cost}, and oracle quantities use the complete
candidate set.
}
\label{tab:stage_ablation}
\end{table*}

\begin{figure}[!t]
\centering
\includegraphics[
width=0.95\columnwidth
]{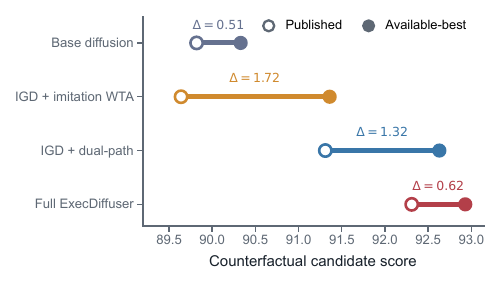}
\caption{
Proposal-only contradiction on Val14. Both proposal generators use
the same imitation publication rule. Stronger proposals raise the
available-best ceiling, but published quality fails to follow.
Values are five-seed means.
}
\label{fig:proposal_contradiction}
\end{figure}

\paragraph{Stronger proposals can widen the gap.}
Figure~\ref{fig:proposal_contradiction} compares the base proposal
generator with the WTA-trained IGD generator under the same imitation
selector. The latter, whose oracle-best quality is 91.36, then forms the
starting proposal configuration in Table~\ref{tab:stage_ablation}(a),
where the full EGS selector is fixed. In the proposal-only comparison,
oracle-best quality rises from 90.33 to 91.36, yet published quality
falls from 89.82 to 89.64, expanding regret from 0.51 to 1.72. Paired
95\% intervals are $+1.03\,[0.76,1.29]$ for oracle-best quality,
$-0.18\,[-0.34,-0.03]$ for published quality, and
$+1.21\,[0.93,1.48]$ for regret. The direction also holds for every
training seed, showing that stronger availability alone does not
guarantee a stronger executed decision.

\paragraph{IGD raises the available-candidate ceiling.}
With EGS fixed, responsibility and shaping raise oracle-best quality
from 91.36 to 92.63, increase the feasible-candidate ratio from
86.60\% to 92.10\%, and reduce all-invalid sets from 2.17\% to 0.82\%
(Table~\ref{tab:stage_ablation}(a)). Specialization further raises the
ceiling to 92.93 and lowers all-invalid sets to 0.68\%. Appendix Sec.~C shows that responsibility and geometry
shaping are independently beneficial, while their dual-path
combination gives the largest candidate-ceiling gain. The gain persists under a matched average decoded budget$\bar K=5.5$ (Appendix Table~A18), ruling out candidate count as the sole explanation.

\begin{figure}[!t]
\centering
\includegraphics[
width=0.95\columnwidth
]{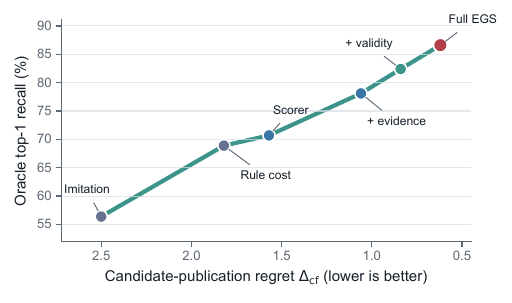}
\caption{
Publication cascade with the proposal generator fixed.
Candidate-conditioned scoring, online evidence, hard validity, and
recurrent context progressively improve oracle top-1 recall and reduce
candidate-publication regret. Values are five-seed means.
}
\label{fig:selection_quality}
\end{figure}

\paragraph{EGS converts available quality into publication quality.}
With the full proposal generator fixed, rule cost improves oracle top-1
recall from 56.4\% to 68.9\% and reduces regret from 2.50 to 1.82,
confirming that online execution evidence is useful. Full EGS reaches
86.6\% recall and 0.62 regret, corresponding to a 30.2-percentage-point
recall gain and a 75.2\% regret reduction over imitation. The remaining
gain indicates that context-dependent publication benefits from learned
candidate utility beyond the tested fixed aggregation
(Table~\ref{tab:stage_ablation}(b) and
Fig.~\ref{fig:selection_quality}). Appendix Sec.~C reports the complete
scorer--evidence--validity--recurrence cascade.

\paragraph{The initial publication changes later replanning.}
Across commit horizons from 0.5 to 8.0~s, EGS reduces regret by
66.9--75.2\% relative to imitation while maintaining 86.1--88.1\%
oracle top-1 recall
(Appendix Sec.~D.1). We then
intervene only on the initial publication: each branch executes a
selector-specific prefix for 0.5--4.0~s and subsequently resumes the
same frozen full \ours planner for the remainder of a common 8.0-s
window. After the forced prefix, every branch shares IGD, EGS,
forecasting, allocation, continuity, validity, and MRM; only the
physically executed initial decision differs.

\begin{table*}[!t]
\centering
\small
\setlength{\tabcolsep}{5.0pt}
\begin{tabular}{@{}lcccc@{}}
\toprule
\shortstack{Prefix\\horizon} &
\shortstack{Unique scenarios /\\eligible states} &
\shortstack{$\Delta$ score vs. imitation\\(paired 95\% CI)} &
\shortstack{$\Delta$ score vs. rule cost\\(paired 95\% CI)} &
\shortstack{Recovery rate (\%)\\imitation / EGS} \\
\midrule
0.5~s
& 1,118 / 111,820
& $+0.51\,[0.33,0.69]$
& $+0.37\,[0.21,0.53]$
& 84.7 / 87.8 \\
1.0~s
& 1,118 / 111,406
& $+0.91\,[0.66,1.15]$
& $+0.62\,[0.39,0.84]$
& 82.9 / 88.3 \\
2.0~s
& 1,117 / 110,736
& $+1.42\,[1.08,1.75]$
& $+0.95\,[0.66,1.23]$
& 79.8 / 87.1 \\
4.0~s
& 1,115 / 109,284
& $+1.90\,[1.44,2.34]$
& $+1.25\,[0.87,1.62]$
& 75.4 / 85.9 \\
\bottomrule
\end{tabular}
\normalsize
\caption{
Resumed replanning after prefix intervention. Score differences
compare EGS with imitation and rule-cost prefixes; intervals are paired
over scenario clusters. Recovery uses the same degraded-state
denominator within each horizon.
}
\label{tab:resumed_main}
\end{table*}

Table~\ref{tab:resumed_main} shows that the resumed-score advantage
over imitation grows from 0.51 to 1.90 points as the prefix lengthens,
with every paired interval excluding zero. Recovery is evaluated on a
common degraded-state set and requires the final 3.0~s to remain
collision- and MRM-free, with valid proposals, sufficient TTC, and
restored route progress. EGS also outperforms rule-cost publication at
every horizon and maintains consistently higher recovery rates. The
growing separation shows that publication changes the state from which
the same planner must continue.

\begin{figure}[!t]
\centering
\includegraphics[
width=\columnwidth
]{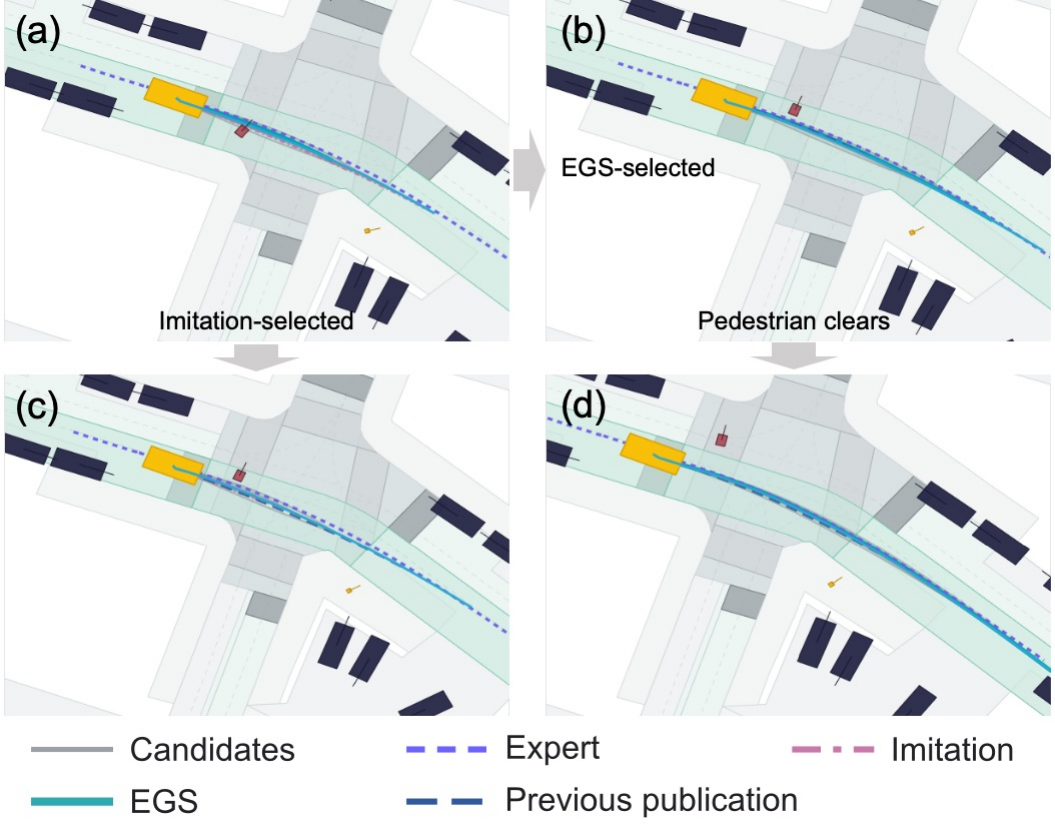}
\caption{
Matched-prefix intervention on nuPlan. Different initial publications
create different physical states before the same frozen planner
resumes.
}
\label{fig:counterfactual_feedback}
\end{figure}

Figure~\ref{fig:counterfactual_feedback} visualizes one matched branch
pair. The effect is smallest in straight following and largest in
interaction-intensive cases: at a 4.0-s prefix, the EGS advantage rises
from 0.39 points in straight following to 2.04 points in unprotected
turns. Complete branch results and recovery definitions appear in
Appendix Tables~A13--A16, and scenario-wise effects in
Fig.~A4.

\subsection{Cross-Benchmark Evaluation on NAVSIM}
\label{sec:navsim_evaluation}

We retrain \ours on NAVSIM~\cite{dauner2024navsim} v1.1 from scratch,
without external data or pretrained checkpoints, and evaluate five
independently trained models using the official \texttt{navtest} PDMS
pipeline.

\begin{table}[!t]
\centering
\small
\setlength{\tabcolsep}{2.0pt}
\begin{tabular}{@{}lccccc@{}}
\toprule
Method & NC & DAC & TTC & EP & PDMS \\
\midrule
DiffusionDrive~\cite{liao2025diffusiondrive}
& 98.2 & 96.2 & 94.7 & 82.2 & 88.1 \\
CdDrive~\cite{wu2026cddrive}
& 98.7 & 97.5 & 95.4 & 82.7 & 89.2 \\
ReCogDrive~\cite{li2026recogdrive}
& 97.9 & 97.3 & 94.9 & 87.3 & 90.8 \\
DiffusionDriveV2~\cite{zou2025diffusiondrivev2}
& 98.3 & 97.9 & 94.8 & 87.5 & 91.2 \\
\midrule
\textbf{\ours}
& \textbf{99.1}
& \textbf{98.2}
& \textbf{97.4}
& \textbf{87.6}
& \textbf{92.6} \\
\bottomrule
\end{tabular}
\normalsize
\caption{
NAVSIM v1.1 \texttt{navtest} results (\%). Baselines are
reported-paper values; \ours is the five-seed mean.
}
\label{tab:navsim_generalization}
\end{table}

\ours achieves 92.6 PDMS, 1.4 points above the reported
DiffusionDriveV2 result, and leads the compared methods on NC, DAC,
TTC, and EP (Table~\ref{tab:navsim_generalization}). The balanced gains
in collision avoidance, interaction safety, and ego progress indicate
that the improvement is not driven by conservative stopping. Across
five seeds, PDMS ranges from 92.2 to 92.9 with a standard deviation of
0.29 (Appendix Table~A23 and Fig.~A10). These results show that
the proposal-to-execution design remains effective beyond the nuPlan
evaluation setting; Appendix Sec.~F provides the complete
protocol and per-seed results.

\begin{figure}[!t]
\centering
\includegraphics[
width=\columnwidth
]{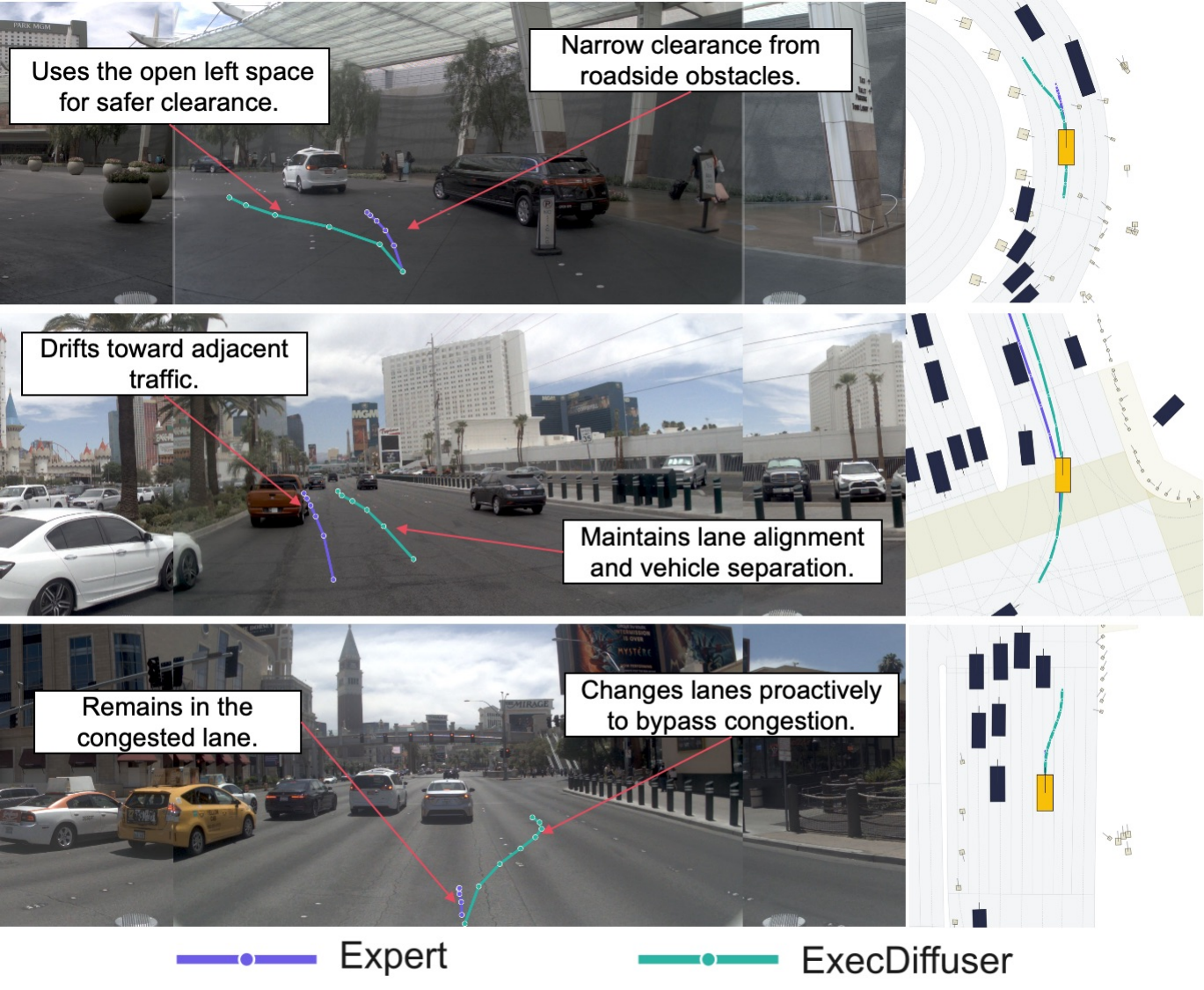}
\caption{
Qualitative NAVSIM examples of an open-space lane change,
lane-aligned motion after a turn, and navigation around queued traffic.
Orange denotes the logged expert reference and cyan the trajectory
selected by \ours.
}
\label{fig:navsim_qualitative}
\end{figure}

\subsection{Reliability, Failure Boundaries, and Cost}
\label{sec:reliability_generalization_cost}

The Adaptive Continuity Gate reduces plan switching from 13.6\% to
7.3\% while limiting hazard-response delay to 0.18~s and achieving
93.4\% hazard success (Appendix Table~A11). Uncertainty-aware
forecasting detects 87.6\% of counterfactually required MRM events,
but the false-valid rate rises from 2.70\% overall to 5.14\% in
occluded intersections, exposing a remaining forecast-dependent
failure boundary (Appendix Fig.~A5 and Tables~A21--A22).

The gain does not arise from conservative slowing: \ours maintains
6.39~m/s average speed, versus 6.42~m/s for Diffusion Planner, while
reducing unnecessary stops, high-jerk events, and emergency braking
(Appendix Fig.~A9). It adds 5.2M parameters and runs at 59.5~ms,
compared with 50.7~ms for Diffusion Planner and 47.9~ms for FlowDrive
(Appendix Table~A4). Adaptive allocation further outperforms
fixed $K=8$ on Val14-R (87.11 vs.\ 86.52) while reducing latency from
80.0 to 59.5~ms (Appendix Fig.~A6(b) and Table~A20).

\section{Conclusion}
\label{sec:conclusion}
We investigated whether improving the trajectories available to a generative planner necessarily improves the trajectory executed by the vehicle. Our
experiments reveal that these objectives are distinct. With the selector fixed, stronger proposal learning increases the quality of the best available
candidate, but can reduce the quality of the committed trajectory and enlarge the candidate-publication regret. This result identifies a previously underemphasized failure mode in diffusion planning: a higher proposal ceiling
does not guarantee a better closed-loop decision.

The controlled ablations further show that the two sides of this failure have different causes. Execution-aware proposal learning improves candidate
availability by increasing the quality and validity of the generated set, whereas execution-grounded selection improves the conversion of available
candidates into committed decisions. With the proposal generator fixed, the learned selector substantially improves oracle-selection recall and reduces candidate-publication regret. These results indicate that trajectory selection should not be treated as a passive post-processing step after generation, but as an execution-critical component of the planner.

The benefit also persists beyond retrospective candidate evaluation. Prefix interventions show that different initial commitments produce different
physical states, after which the same frozen planner exhibits different replanning performance and recovery behavior. Across nuPlan, InterPlan, and NAVSIM, \ours achieves strong closed-loop results, with particularly clear gains under reactive and interaction heavy conditions. Overall, our findings suggest that generative planners should be evaluated along two complementary dimensions: what they can
generate and what they ultimately commit to control. Future work should address forecast uncertainty, false-valid decisions in occluded scenes, and
validation in real-world interactive environments.

\clearpage
\appendices
\setcounter{section}{0}
\setcounter{figure}{0}
\setcounter{table}{0}
\setcounter{equation}{0}
\renewcommand{\thesection}{\Alph{section}}
\renewcommand{\thesubsection}{\thesection.\arabic{subsection}}
\renewcommand{\thefigure}{A\arabic{figure}}
\renewcommand{\thetable}{A\arabic{table}}
\renewcommand{\theequation}{A\arabic{equation}}
\section{Protocol and Reproducibility}
\label{app:protocol}

\noindent
These appendices provide the complete implementation,
statistical, ablation, intervention, reliability, and NAVSIM records
supporting the main paper. Appendices~\ref{app:protocol}--\ref{sec:app_navsim}
follow the same evidence chain as the main paper.

This section specifies the implementation and fixed calibration used
across all seeds and controlled ablations.

\subsection{Training and Evaluation Configuration}

\begin{table*}[!t]
\centering
\small
\begin{tabular}{lc}
\toprule
Configuration & Value \\
\midrule
Training data & nuPlan v1.1 training split \\
Calibration split & Scenario-disjoint 10\% of training data \\
Observation / planning horizon & 2.0~s / 8.0~s \\
Planning frequency & 10~Hz \\
Intent queries / budgets & $8$ / $\{2,4,8\}$ \\
Training / inference diffusion steps & 1000 / 25 \\
Optimizer / learning rate & AdamW / $5\times10^{-4}$ \\
Per-GPU / effective batch size & 128 / 1024 \\
Training epochs & 500 \\
Training hardware & $8\times$ NVIDIA GeForce RTX~4090D (24~GB each) \\
Runtime hardware & $1\times$ NVIDIA GeForce RTX~4090D, batch 1 \\
CPU & $2\times$ Intel Xeon Silver 4416+ \\
CPU cores / threads & 40 physical cores / 80 threads total \\
System memory & 256~GB DDR5 \\
Software & Ubuntu 22.04; Python 3.9 \\
Framework & PyTorch 2.0.0+cu118; CUDA 11.8 \\
Random seeds & $\{0,1,2,3,4\}$ \\
Main-table reporting & Arithmetic mean over all five seeds \\
Seed selection / ensembling & None / none \\
Latency protocol & 200 warm-up + 1,000 timed calls \\
\bottomrule
\end{tabular}
\normalsize
\caption{Training, timing, and reporting configuration.}
\label{tab:implementation_details}
\end{table*}

Every candidate and method pair uses identical scenario seeds,
simulator settings, and metric implementations. Calibration statistics,
thresholds, and loss coefficients are fixed before all benchmark runs.

\subsection{Candidate Representation, Recurrent Alignment, and Gradient Routing}
\label{app:candidate_representation}

Each active intent query $q_m$ is paired with one independently sampled
diffusion noise variable. The decoder predicts a normalized clean state
sequence $\hat{\mathbf x}_{0,t}^{(m)}$. A fixed differentiable operator
$\Pi$ denormalizes the sequence and reconstructs position, heading,
velocity, acceleration, and other trajectory states:
\begin{equation}
\tau_t^{(m)}
=
\Pi\!\left(\hat{\mathbf x}_{0,t}^{(m)}\right).
\label{eq:app_trajectory_recovery}
\end{equation}
The training preference cost, online evidence, hard validity, and
differentiable shaping objective are all evaluated on this physical trajectory
representation. This avoids defining proposal responsibility on one
trajectory representation while optimizing or ranking another.

The trajectory published at cycle $t-1$ is converted to the current
ego frame by
\begin{equation}
\bar{\tau}_{t-1}^{\mathrm{pub}}
=
\operatorname{Align}
\left(
\tau_{t-1}^{\mathrm{pub}},
s_t^{\mathrm{ego}}
\right).
\label{eq:app_plan_alignment}
\end{equation}
The alignment removes the executed prefix, applies the relative SE(2)
transform, and resamples the unexecuted suffix to the current planning
grid. The stored trajectory is treated as recurrent state and detached
across planning cycles; no gradient is propagated through the simulator
or through a previous publication decision.

For EGS training, sampled candidate coordinates are detached before the
trajectory embedding and candidate-conditioned attention:
\begin{equation}
\begin{aligned}
\mathbf z_t^{(k)}
&=\operatorname{Concat}\big[
f_\tau(\operatorname{sg}[\tau_t^{(k)}]),
f_{\mathrm{attn}}(\operatorname{sg}[\tau_t^{(k)}],E_t),\\
&\hspace{1.2em}
f_{\mathrm{prev}}(\bar{\tau}_{t-1}^{\mathrm{pub}}),
f_e(\mathbf e_t^{(k)}),a_t\big].
\end{aligned}
\label{eq:app_ranking_detach}
\end{equation}
Route context enters through $e_{\mathrm{route},t}^{(k)}$, and forecast
context enters through risk evidence computed from predicted agent
futures. Consequently, $\mathcal L_{\mathrm{rank}}$ does not
backpropagate through candidate coordinates or the diffusion decoding
path. The scorer and non-detached contextual representations may still
receive ranking gradients. Direct coordinate-level proposal optimization
is supplied only by $\mathcal L_{\mathrm{shape}}$.

\paragraph{Training preference, online rule cost, and learned utility.}
The execution-conditioned training preference
$\mathcal A_{\mathrm{train},t}^{(k)}$ uses the same normalized online
evidence as the rule control, then adds ground-truth imitation proximity
and the hard-invalid penalty. Its detached distribution
$p_{\mathrm{train},t}^{(k)}$ supervises hypothesis ownership and
publication learning and is available only during training. The
rule-cost control instead ranks admitted candidates with the online
analytic evidence cost
\begin{equation}
\mathcal C_{\mathrm{rule},t}^{(k)}
=
\lambda_{\mathrm{exec}}\mathbf w^\top
\mathbf e_{t,\mathrm{nt}}^{(k)}
+
\lambda_{\mathrm{temp}}a_t e_{\mathrm{temp},t}^{(k)}.
\label{eq:app_online_rule_cost}
\end{equation}
Full EGS does not add either cost to its inference score. It infers a
context-dependent utility $u_\phi^{(k)}$ from online candidate geometry,
scene tokens, route and forecast evidence, and recurrent context; unlike
$\mathcal C_{\mathrm{rule},t}^{(k)}$, this utility is not restricted to
a fixed linear evidence aggregation. Consequently,
$\mathcal L_{\mathrm{resp}}$ redistributes supervision across
query--noise hypotheses, $\mathcal L_{\mathrm{shape}}$ directly updates
proposal coordinates through $\Pi$, and $\mathcal L_{\mathrm{rank}}$
trains the utility used for publication.

\subsection{Counterfactual Evaluator and Diagnostic State Set}
\label{app:counterfactual_protocol}

A diagnostic planning state is sampled from the normal closed-loop run.
The simulator, ego state, agent state, map, route, and random state are
cloned. Each generated candidate is then committed for 8.0~s without
replanning. Under reactive evaluation, the reactive-agent model is
reinitialized for every candidate, so each rollout starts from the same
state but responds independently to the committed ego trajectory.
Candidate scores use the same official collision, TTC, drivable-area,
direction, speed, comfort, and progress implementations and the same
aggregation order for selected and oracle-best candidates.
Oracle-best is evaluated over the complete generated set
$\{\tau_i^{(k)}\}_{k=1}^{K_i}$, including candidates rejected by the
online hard-validity mask. Hard validity restricts publication but does
not restrict the retrospective oracle.

\paragraph{Diagnostic-state construction and pooling.}
The diagnostic evaluator is executed at every 10-Hz planning cycle
for which the required simulator state and rollout fields are
available; no temporal subsampling is applied. We exclude
post-termination cycles and states with an incomplete candidate set,
hard-validity mask, counterfactual rollout output, or evaluation
horizon. A terminal cycle is retained only when the complete
diagnostic horizon is available.

Let $\mathcal C_i=\{\tau_i^{(k)}\}_{k=1}^{K_i}$ denote the generated
candidate set,
$\mathbf m_i^{\mathrm{hard}}
=\{m_i^{(k)}\}_{k=1}^{K_i}$ the hard-validity mask, and
$\mathbf y_i^{\mathrm{cf}}
=\{S_{\mathrm{cf}}(\tau_i^{(k)})\}_{k=1}^{K_i}$ the complete
candidate-level counterfactual scores. Define
\begin{equation}
\begin{aligned}
c_i^{\mathrm{diag}}
&=
\mathbb I\!\left[
\begin{array}{l}
\mathcal C_i,\,
\mathbf m_i^{\mathrm{hard}},\,
\mathbf y_i^{\mathrm{cf}}\\
\text{and the required horizon are complete}
\end{array}
\right],
\\
\Omega_{\mathrm{diag}}
&=
\{i:c_i^{\mathrm{diag}}=1\}.
\end{aligned}
\label{eq:app_diagnostic_state_set}
\end{equation}
Availability of a previous publication is required only for
temporal-feedback diagnostics and is not part of this base eligibility
criterion. With $p_{i-1}$ denoting the preceding publication and
\begin{equation}
n_i^{\mathrm{valid}}
=
\sum_{k=1}^{K_i}m_i^{(k)},
\end{equation}
we define
\begin{align}
\Omega_{\mathrm{temp}}
&=
\{i\in\Omega_{\mathrm{diag}}:p_{i-1}\text{ is available}\},\\
\Omega_{\mathrm{sel}}
&=
\{i\in\Omega_{\mathrm{diag}}:n_i^{\mathrm{valid}}\geq1\},\\
\Omega_{\mathrm{APD}}
&=
\{i\in\Omega_{\mathrm{diag}}:n_i^{\mathrm{valid}}\geq2\}.
\label{eq:app_selection_state_set}
\end{align}
Temporal-feedback diagnostics use $\Omega_{\mathrm{temp}}$.
Candidate-publication regret and oracle top-1 recall use
$\Omega_{\mathrm{sel}}$. All-invalid states that invoke MRM form
$\Omega_{\mathrm{diag}}\setminus\Omega_{\mathrm{sel}}$ and are
reported separately with $|\Omega_{\mathrm{diag}}|$ as the
denominator.

Across seeds 0--4, $\Omega_{\mathrm{diag}}$ contains 22,421, 22,586,
22,711, 22,494, and 22,628 eligible states, respectively, for a total
of $|\Omega_{\mathrm{diag}}|=112{,}840$. The counts differ slightly
because closed-loop rollouts may terminate at different frames and
therefore produce different numbers of diagnostically complete
states. All seeds use the same scenario filter and evaluator
configuration, and no scenario-level evaluation jobs are missing.
Unless stated otherwise, planning-state rates pool event and eligible
state counts over all seeds before taking their ratio, so every
eligible planning state receives equal weight.

The candidate-publication regret is
\begin{equation}
\Delta_{\mathrm{cf}}
=
\frac{1}{|\Omega_{\mathrm{sel}}|}
\sum_{i\in\Omega_{\mathrm{sel}}}
\left[
\max_{1\leq k\leq K_i}S_{\mathrm{cf}}(\tau_i^{(k)})
-
S_{\mathrm{cf}}(\tau_i^{\mathrm{pub}})
\right].
\label{eq:app_counterfactual_regret}
\end{equation}
Accordingly, $\Delta_{\mathrm{cf}}$ isolates the publication side of
the sample-to-execution gap over the complete generated set. It includes
ranking error among admitted candidates and false rejection when hard
validity removes the all-candidate oracle. Its downstream effect is
evaluated separately through forced-prefix execution and resumed
replanning. Let
$\mathcal O_i=\arg\max_k S_{\mathrm{cf}}(\tau_i^{(k)})$.
Oracle top-1 recall counts a hit whenever the published candidate belongs
to $\mathcal O_i$, so tied oracle scores are handled without arbitrary
tie-breaking. A retrospectively optimal candidate rejected by hard
validity counts as a miss. All-invalid
states that invoke MRM are excluded from candidate-publication regret and
top-1 recall, and are reported separately as proposal-failure/MRM
events. The evaluator is not used for training, calibration, checkpoint
selection, allocation targets, continuity labels, EGS scoring, or online
inference, and it shares no learned parameters with the forecast head or
scorer.

\paragraph{Statistical units.}
Training variability is summarized across independently trained
models. Confidence intervals for method comparisons use 10,000 paired
bootstrap resamples at the scenario level. Each replicate samples
scenarios with replacement while preserving all associated method and
seed results. Individual planning states are never treated as
independent bootstrap units.

\subsection{Execution Evidence, Normalization, and Hard Validity}
\label{app:auxiliary_objectives}

For each evidence dimension $d$, we construct a nonnegative raw cost
$r_{t,d}^{(k)}$ and apply a monotone robust normalizer calibrated on the
scenario-disjoint split:
\begin{equation}
e_{t,d}^{(k)}
=
\operatorname{Norm}_d
\left(r_{t,d}^{(k)}\right).
\label{eq:app_evidence_normalization}
\end{equation}
Normalization preserves the common orientation $0$ = favorable and
larger = worse.

\begin{table*}[!t]
\centering
\small
\setlength{\tabcolsep}{4pt}
\begin{tabular}{p{0.15\textwidth}p{0.57\textwidth}p{0.20\textwidth}}
\toprule
Evidence & Nonnegative raw cost & Direction \\
\midrule
$e_{\mathrm{drive}}$ & Swept-footprint fraction outside drivable polygons & More outside area $\uparrow$ cost \\
$e_{\mathrm{route}}$ & Distance/area outside the commanded route corridor & More route deviation $\uparrow$ cost \\
$e_{\mathrm{lane}}$ & Lane-boundary overlap and lane-topology inconsistency & More violation $\uparrow$ cost \\
$e_{\mathrm{risk}}$ & Predicted overlap plus clearance and TTC shortfalls & More overlap/shortfall $\uparrow$ cost \\
$e_{\mathrm{progress}}$ & Shortfall from route-aligned progress target & Less progress $\uparrow$ cost \\
$e_{\mathrm{kin}}$ & Speed, acceleration, curvature, and yaw-rate exceedance & More exceedance $\uparrow$ cost \\
$e_{\mathrm{comfort}}$ & Lateral-acceleration and jerk exceedance & More exceedance $\uparrow$ cost \\
$e_{\mathrm{temp}}$ & Position, heading, velocity, and acceleration deviation from the aligned previous publication & More deviation $\uparrow$ cost \\
\bottomrule
\end{tabular}
\normalsize
\caption{Orientation of the online execution evidence. These normalized
quantities enter the training preference, rule-cost control, and learned
evidence encoder with the same penalty direction. Coordinate shaping uses
aligned differentiable surrogates defined in
Eq.~\eqref{eq:app_shape_objective}.}
\label{tab:evidence_orientation}
\end{table*}

The hard-validity mask is the conjunction of drivable-region and imminent
collision checks. Drivable validity tests the swept ego footprint and
ego-center tolerance. Collision validity uses uncertainty-inflated predicted
agent footprints over the fixed look-ahead horizon and rejects a candidate
when the swept overlap or TTC condition crosses the calibrated threshold.
Route deviation, lane consistency, progress, comfort, and moderate risk
remain soft evidence and are not hard rejection rules.

Raw APD is computed over the complete generated set. Feasible APD is
computed only from hard-valid candidates. For each
$i\in\Omega_{\mathrm{APD}}$, let
\begin{equation}
\mathcal V_i
=
\{\tau_i^{(k)}:m_i^{(k)}=1\},
\qquad
n_i^{\mathrm{valid}}=|\mathcal V_i|.
\end{equation}
The distance between two candidates is the mean Euclidean distance
between their time-aligned two-dimensional positions over the
prediction horizon:
\begin{equation}
d(\tau_i^{(a)},\tau_i^{(b)})
=
\frac{1}{T}
\sum_{t=1}^{T}
\left\|
\tau_{i,xy}^{(a)}(t)-\tau_{i,xy}^{(b)}(t)
\right\|_2 ,
\label{eq:app_trajectory_distance}
\end{equation}
which is measured in meters and does not use heading, velocity,
learned EGS score, or validity cost. The state-level feasible APD is
\begin{equation}
\mathrm{APD}_i
=
\frac{2}
{n_i^{\mathrm{valid}}(n_i^{\mathrm{valid}}-1)}
\sum_{\substack{a<b\\
\tau_i^{(a)},\tau_i^{(b)}\in\mathcal V_i}}
d(\tau_i^{(a)},\tau_i^{(b)}),
\end{equation}
and the reported value is
\begin{equation}
\mathrm{Feasible\ APD}
=
\frac{1}{|\Omega_{\mathrm{APD}}|}
\sum_{i\in\Omega_{\mathrm{APD}}}\mathrm{APD}_i.
\end{equation}
States with fewer than two hard-valid candidates are excluded from
$\Omega_{\mathrm{APD}}$ and are not assigned zero diversity. In our
diagnostics, $|\Omega_{\mathrm{APD}}|=109{,}376$ of the
$|\Omega_{\mathrm{diag}}|=112{,}840$ states satisfy this criterion,
giving
\begin{equation}
\mathrm{Coverage}_{\mathrm{APD}}
=
\frac{|\Omega_{\mathrm{APD}}|}
{|\Omega_{\mathrm{diag}}|}
=
\frac{109{,}376}{112{,}840}
=
96.93\%.
\label{eq:app_feasible_apd_coverage}
\end{equation}
The remaining 3,464 states are excluded from the feasible-APD
denominator. Feasible APD is interpreted jointly with this coverage,
the feasible-candidate ratio, and the all-invalid-set rate.

\subsection{Adaptive Continuity Gate}

The Adaptive Continuity Gate uses the current scene representation,
scene change, and current evidence of the aligned previous publication:
\begin{equation}
a_t
=
\sigma
\left(
g_\chi
\left[
h_t,\,
h_t-h_{t-1},\,
\mathbf e_t^{\mathrm{prev}}
\right]
\right).
\label{eq:app_continuity_gate}
\end{equation}
Let $m_t^{\mathrm{prev}}$ be the hard-validity result for the aligned
previous publication and let $e_{\mathrm{risk},t}^{\mathrm{prev}}$ be its
normalized risk cost. The training target is
\begin{equation}
y_t^{\mathrm{cont}}
=
\mathbb I
\left[
m_t^{\mathrm{prev}}=1
\;\land\;
e_{\mathrm{risk},t}^{\mathrm{prev}}
\leq\tau_{\mathrm{cont}}
\right],
\label{eq:app_gate_target}
\end{equation}
with binary cross-entropy
\begin{equation}
\mathcal L_{\mathrm{gate}}
=
-\frac{1}{B}
\sum_{b=1}^{B}
\left[
y_b^{\mathrm{cont}}\log a_b
+
(1-y_b^{\mathrm{cont}})
\log(1-a_b)
\right].
\label{eq:app_gate_loss}
\end{equation}
The gate modulates temporal evidence but never overrides hard validity.

\subsection{Shared Shaping Weights and Query Specialization}
\label{app:shaping_specialization}

The normalized online evidence and training preference share a common
penalty direction and relative weights. Coordinate correction uses a
separate differentiable surrogate objective while preserving these
semantics. For
\[
\mathcal D_{\mathrm{diff}}
=
\{
\mathrm{route},
\mathrm{lane},
\mathrm{risk},
\mathrm{progress},
\mathrm{kin},
\mathrm{comfort}
\},
\]
we set
\begin{equation}
\begin{aligned}
\alpha_d&=w_d,
&\alpha_{\mathrm{temp}}
&=\frac{\lambda_{\mathrm{temp}}}{\lambda_{\mathrm{exec}}},\\
\bar\alpha_d
&=\frac{\alpha_d}
{\sum_{j\in\mathcal D_{\mathrm{diff}}\cup\{\mathrm{temp}\}}\alpha_j}.
\end{aligned}
\label{eq:app_shape_weights}
\end{equation}
Let $\tilde e_{t,d}^{(k)}$ denote a smooth, coordinate-differentiable
surrogate with the same penalty orientation as $e_{t,d}^{(k)}$. The
shaping objective is
\begin{equation}
\widetilde{\mathcal C}_{\mathrm{shape}}(\tau_t^{(k)})
=
\sum_{d\in\mathcal D_{\mathrm{diff}}}
\bar\alpha_d\tilde e_{t,d}^{(k)}
+
\bar\alpha_{\mathrm{temp}}a_t
\tilde e_{\mathrm{temp},t}^{(k)}.
\label{eq:app_shape_objective}
\end{equation}
The surrogate quantities in Eq.~\eqref{eq:app_shape_objective} are used
only for coordinate-level shaping; they do not replace the normalized
online evidence in the training preference, rule-cost baseline, or EGS
input. The global coefficient $\lambda_{\mathrm{shape}}$ controls the
strength of this path. Discrete collision and off-road indicators are
excluded from the differentiable set and remain detached.

Let $\mu(k)$ identify the query that generated candidate $k$. The
minibatch-level responsibility mass of query $m$ is
\begin{equation}
\bar p_m
\!=\frac{1}{B}\sum_{b=1}^{B}
\sum_{k:\mu(k)=m}p_{\mathrm{train},b}^{(k)},
\qquad
\bar{\mathbf p}
\!=[\bar p_1,\ldots,\bar p_M].
\label{eq:app_query_mass}
\end{equation}
We use
\begin{equation}
\mathcal L_{\mathrm{spec}}
=
D_{\mathrm{KL}}
\left(
\bar{\mathbf p}\,\|\,\mathbf u_M
\right)
+
\lambda_{\mathrm{sep}}
\mathcal L_{\mathrm{sep}},
\label{eq:app_specialization_loss}
\end{equation}
where $\mathbf u_M$ is uniform and
$\mathcal L_{\mathrm{sep}}$ penalizes insufficient separation between
route-valid query outputs. This term supports coverage but does not
replace execution-conditioned responsibility.

\subsection{Scene-Conditioned Candidate Allocation}
\label{app:allocation_objective}

The allocation head predicts a budget distribution and one pre-decoding
priority score per intent query:
\begin{equation}
\left[p_\omega(K_t\mid h_t,c_t),
\{s_t^{(m)}\}_{m=1}^{M}\right]
=A_\omega\!\left(h_t,c_t,\mathcal Q\right).
\label{eq:app_allocation_head}
\end{equation}
The priority head uses only $h_t$, $c_t$, and $q_m$; it does not use a
decoded candidate, candidate evidence, or an EGS score. It can therefore
select the top-$K_t$ queries before diffusion decoding and provides a real
compute reduction.

After proposal warm-up, the full $M$-query set is decoded during training
to construct training targets. Candidate-level responsibilities are
aggregated to queries:
\begin{equation}
r_t^{(m)}
\!=\sum_{k:\mu(k)=m}p_{\mathrm{train},t}^{(k)},
\qquad
\mathbf r_t
\!=[r_t^{(1)},\ldots,r_t^{(M)}].
\label{eq:app_priority_target}
\end{equation}
Because each active query produces one candidate, the aggregation reduces
to the corresponding candidate responsibility, but the notation also
covers grouped query--noise variants. The predicted priority distribution
is
\begin{equation}
\begin{aligned}
\pi_t^{(m)}
&=\frac{\exp(s_t^{(m)}/\tau_p)}
{\sum_j\exp(s_t^{(j)}/\tau_p)},\\
\mathcal L_{\mathrm{pri}}
&=D_{\mathrm{KL}}\!\left(
\operatorname{sg}[\mathbf r_t]\,\|\,\boldsymbol\pi_t
\right).
\end{aligned}
\label{eq:app_priority_loss}
\end{equation}

The training-preference order sorts queries by decreasing $r_t^{(m)}$. Let
$\mathcal T_t^\dagger(K)$ denote its first $K$ candidates. Because a lower training preference cost is better, define
\begin{equation}
C_t(K)
=
\min_{k\in\mathcal T_t^\dagger(K):\,m_t^{(k)}=1}
\mathcal A_{\mathrm{train},t}^{(k)},
\label{eq:budget_best_cost}
\end{equation}
with $C_t(K)=+\infty$ when the prefix contains no hard-valid proposal.
The admissible budget set is
\begin{equation}
\mathcal K_{\epsilon,t}
=
\left\{
K\in\mathcal K
\;\middle|\;
C_t(K)\leq C_t(K_{\max})+\epsilon_K
\right\}.
\label{eq:budget_admissible_set}
\end{equation}

The budget target is
\begin{equation}
K_t^\dagger
=
\begin{cases}
\min \mathcal K_{\epsilon,t},
& C_t(K_{\max})<+\infty,\\[3pt]
K_{\max},
& C_t(K_{\max})=+\infty.
\end{cases}
\label{eq:budget_target}
\end{equation}
The budget loss is
\begin{equation}
\begin{aligned}
\mathcal L_K
&\!=-\frac{1}{B}\sum_{b=1}^{B}
\log p_\omega\!\left(K_b^\dagger\mid h_b,c_b\right),
\\
\mathcal L_{\mathrm{alloc}}
&\!=\mathcal L_K+\mathcal L_{\mathrm{pri}}.
\end{aligned}
\label{eq:allocation_loss}
\end{equation}
A smaller budget is selected only when the training-prioritized prefix
preserves the best full-budget hard-valid cost within $\epsilon_K$.
At inference, the learned scene-conditioned order replaces the training-preference
order, and only its top-$K_t$ queries are decoded.

\subsection{Forecast Supervision}

Let $\hat{\mathbf s}_{i,r}$ and
$\mathbf s_{i,r}^{\mathrm{gt}}$ be the predicted and ground-truth
normalized states of agent $i$ at future step $r$, and let $v_{i,r}$
mask unavailable states. We use
\begin{equation}
\mathcal L_{\mathrm{pred}}
=
\frac{
\sum_i\sum_{r=1}^{T}
v_{i,r}
\rho_{\mathrm{Huber}}
\left(
\hat{\mathbf s}_{i,r}
-
\mathbf s_{i,r}^{\mathrm{gt}}
\right)
}{
\sum_i\sum_{r=1}^{T}v_{i,r}+\varepsilon
}.
\label{eq:prediction_loss}
\end{equation}
Future states supervise only the training-time forecast head. Online
planning uses predictions produced from observed history and never accesses
ground-truth future agent states.

\subsection{Fixed Verification, Training, Allocation, and MRM Parameters}

Table~\ref{tab:fixed_parameters} lists the fixed verification thresholds,
loss coefficients, allocation settings, continuity parameters, and MRM
constants used throughout the reported experiments.

\begin{table*}[!t]
\centering
\small
\setlength{\tabcolsep}{4pt}
\resizebox{\textwidth}{!}{%
\begin{tabular}{lll}
\toprule
Group & Parameter & Fixed value \\
\midrule
Hard validity & Look-ahead / outside-footprint ratio & 3.0~s / $5\%$ \\
Drivable validity & Ego-center tolerance & 0.30~m \\
Collision validity & Swept overlap / ego inflation & 0.05~m$^2$ / 0.15~m per side \\
TTC validity & Threshold / minimum closing speed & 1.0~s / 0.5~m/s \\
Uncertainty inflation & Radius & $\min(0.80,0.15+0.50\sigma_t)$~m \\
Training-preference weights & drive/route/lane/risk/prog/kin/comfort & $1.00/0.80/0.40/1.50/0.30/0.25/0.20$ \\
Training-preference cost & $\lambda_{\mathrm{exec}}/\lambda_{\mathrm{temp}}/\lambda_{\mathrm{invalid}}$ & $1.00/0.30/20.0$ \\
Core losses & $\lambda_{\mathrm{shape}}/\lambda_{\mathrm{rank}}/\lambda_{\mathrm{spec}}$ & $0.25/1.00/0.10$ \\
Auxiliary losses & $\lambda_{\mathrm{alloc}}/\lambda_{\mathrm{gate}}/\lambda_{\mathrm{pred}}/\lambda_{\mathrm{sep}}$ & $0.05/0.10/0.50/0.20$ \\
Shaping weights & Per-dimension weights & Preference-relative, renormalized over differentiable dimensions \\
Candidate priority & Inputs / application & $(h_t,c_t,q_m)$ / before diffusion decoding \\
Continuity gate & $\tau_{\mathrm{cont}}$ / diagnostic threshold & $0.35/0.50$ \\
Allocation target & Warm-up / tolerance $\epsilon_K$ & 20 epochs / 0.20 score points \\
MRM & Target/max decel / max jerk & $-3.0/-6.0$~m/s$^2$ / 3.0~m/s$^3$ \\
MRM & Braking/full horizon / stop buffer & 4.0/8.0~s / 2.0~m \\
MRM & Curvature / corridor margin / lane change & 0.20~m$^{-1}$ / 0.30~m / disabled \\
\bottomrule
\end{tabular}%
}
\normalsize
\caption{Fixed online-validity, training, allocation, gate, and MRM
parameters.}
\label{tab:fixed_parameters}
\end{table*}

\subsection{Baseline Reproduction and Compute Cost}

\begin{table*}[!t]
\centering
\small
\setlength{\tabcolsep}{3.5pt}
\resizebox{\textwidth}{!}{%
\begin{tabular}{lccccccc}
\toprule
Method & Params (M) & Added params & FLOPs/cycle (G) & Train h/seed & GPU-h/seed & Five-seed GPU-h & Latency (ms) \\
\midrule
Diffusion Planner & 43.7 & -- & 111.8 & 42.6 & 340.8 & 1704.0 & 50.7 \\
FlowDrive & 46.2 & -- & 96.4 & 39.1 & 312.8 & 1564.0 & 47.9 \\
\ours & 48.9 & +5.2 & 138.6 & 49.8 & 398.4 & 1992.0 & 59.5 \\
\bottomrule
\end{tabular}%
}
\normalsize
\caption{Model size and computation under the same NVIDIA GeForce RTX~4090D
environment. FLOPs are measured for batch size one at each method's
default operating point; training cost is reported per seed and over all
five seeds.}
\label{tab:compute_cost}
\end{table*}

The added parameters are concentrated in the shared forecast head, EGS
scorer, continuity gate, and joint allocation head; the scene encoder and
diffusion decoder remain shared. All reproduced methods use the same nuPlan
version, scenario list, metric configuration, five training seeds, and no
post-hoc trajectory refinement. Methods listed only with original-paper
results could not be reproduced under this protocol because their training
implementations are not publicly available; we therefore use them only for
reported-result context, not for controlled improvement claims.

\section{Main-Table Results and Statistical Reliability}
\label{app:statistics}

This section supports the main nuPlan and InterPlan comparison and the
proposal-only contradiction. We report every independently trained model,
sample standard deviations across seeds, and paired scenario-level
intervals for the matched comparisons.

\begin{table*}[!t]
\centering
\small
\setlength{\tabcolsep}{3.2pt}
\begin{tabular}{lccccccc}
\toprule
Seed & Val14 NR & Val14 R & Test14-hard NR & Test14-hard R & Test14 NR & Test14 R & InterPlan \\
\midrule
0 & 92.25 & 86.82 & 79.31 & 74.30 & 92.32 & 88.08 & 38.68 \\
1 & 92.54 & 87.32 & 78.91 & 74.83 & 92.05 & 88.45 & 37.93 \\
2 & 92.34 & 87.17 & 79.18 & 74.57 & 92.21 & 88.41 & 38.46 \\
3 & 92.44 & 87.00 & 79.04 & 74.90 & 92.02 & 88.27 & 38.62 \\
4 & 92.28 & 87.24 & 79.21 & 74.45 & 92.20 & 88.39 & 38.06 \\
\midrule
Mean & 92.37 & 87.11 & 79.13 & 74.61 & 92.16 & 88.32 & 38.35 \\
Std. & 0.12 & 0.20 & 0.16 & 0.25 & 0.12 & 0.15 & 0.34 \\
\bottomrule
\end{tabular}
\normalsize
\caption{Per-seed closed-loop results. The mean row exactly matches
the main paper's closed-loop results table; Std. is the sample standard deviation across
five independently trained models.}
\label{tab:five_seed_results}
\end{table*}

\begin{figure*}[!t]
\centering
\includegraphics[width=0.98\textwidth]{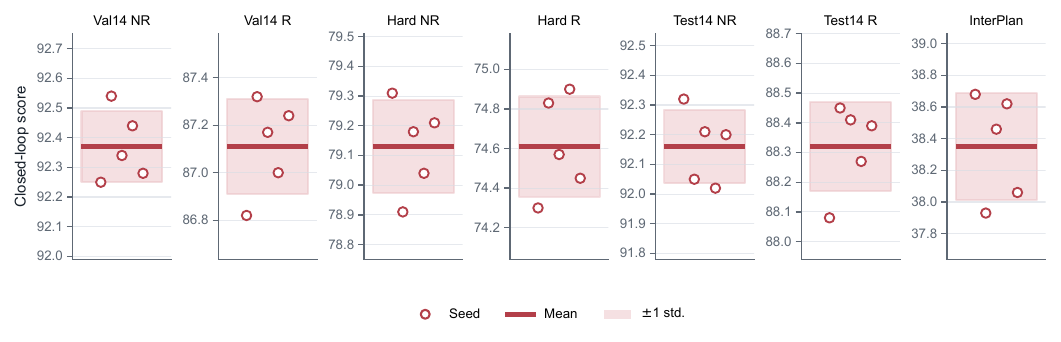}
\caption{Seed-level stability across the seven reported settings. Open circles denote independently trained models, horizontal lines denote arithmetic means, and shaded bands denote one sample standard deviation across the five seeds. Exact values appear in Table~\ref{tab:five_seed_results}.}
\label{fig:appendix_seed_stability}
\end{figure*}

\begin{table*}[!t]
\centering
\small
\begin{tabular}{lcc}
\toprule
Comparison & Difference & Paired 95\% CI \\
\midrule
\ours $-$ Diffusion Planner, Val14 NR & +2.50 & [1.76, 3.21] \\
\ours $-$ Diffusion Planner, Val14 R & +4.73 & [3.68, 5.77] \\
\ours $-$ FlowDrive, Val14 NR & +1.16 & [0.44, 1.86] \\
\ours $-$ FlowDrive, Val14 R & +1.74 & [0.73, 2.71] \\
\ours $-$ Diffusion Planner, Hard R & +5.39 & [4.28, 6.42] \\
\ours $-$ FlowDrive, Hard R & +1.52 & [0.57, 2.46] \\
\bottomrule
\end{tabular}
\normalsize
\caption{Scenario-level paired bootstrap with 10,000 resamples and identical scenario seeds.}
\label{tab:bootstrap_ci}
\end{table*}

\subsection{Five-Seed Consistency of the Proposal-Only Contradiction}
\label{app:proposal_only_consistency}

The central contradiction in the main paper compares the base proposal
model with proposal-only IGD while retaining the same imitation
publication rule. Thus, the proposal set changes but the publication
interface does not. Table~\ref{tab:proposal_only_seed_consistency}
reports every seed. Proposal-only IGD raises oracle-best quality for
all five independently trained models, whereas published quality
slightly decreases for every seed. The resulting regret increase is
therefore not caused by one favorable or unfavorable run.

\begin{table*}[!t]
\centering
\small
\setlength{\tabcolsep}{2.4pt}
\begin{tabular}{@{}lrrrrrrrrr@{}}
\toprule
& \multicolumn{3}{c}{Oracle-best}
& \multicolumn{3}{c}{Published score}
& \multicolumn{3}{c}{Regret} \\
\cmidrule(lr){2-4}
\cmidrule(lr){5-7}
\cmidrule(lr){8-10}
Seed
& Base & IGD & Diff.
& Base & IGD & Diff.
& Base & IGD & Diff. \\
\midrule
0 & 90.28 & 91.29 & $+1.01$ & 89.77 & 89.56 & $-0.21$ & 0.51 & 1.73 & $+1.22$ \\
1 & 90.41 & 91.44 & $+1.03$ & 89.90 & 89.72 & $-0.18$ & 0.51 & 1.72 & $+1.21$ \\
2 & 90.24 & 91.33 & $+1.09$ & 89.75 & 89.59 & $-0.16$ & 0.49 & 1.74 & $+1.25$ \\
3 & 90.35 & 91.39 & $+1.04$ & 89.84 & 89.66 & $-0.18$ & 0.51 & 1.73 & $+1.22$ \\
4 & 90.37 & 91.35 & $+0.98$ & 89.84 & 89.67 & $-0.17$ & 0.53 & 1.68 & $+1.15$ \\
\midrule
Mean & 90.33 & 91.36 & $+1.03$ & 89.82 & 89.64 & $-0.18$ & 0.51 & 1.72 & $+1.21$ \\
\bottomrule
\end{tabular}
\normalsize
\caption{Five-seed consistency of the proposal-only contradiction.
Both variants use the same imitation selector. Diff. is
proposal-only IGD minus Base. The available-best ceiling improves for
every seed, while published quality slightly decreases and regret
increases for every seed.}
\label{tab:proposal_only_seed_consistency}
\end{table*}

\begin{table}[!t]
\centering
\small
\setlength{\tabcolsep}{3.8pt}
\begin{tabular}{lcc}
\toprule
Comparison & Mean diff. & Paired 95\% CI \\
\midrule
Oracle-best & $+1.03$ & $[0.76,1.29]$ \\
Published score & $-0.18$ & $[-0.34,-0.03]$ \\
Regret & $+1.21$ & $[0.93,1.48]$ \\
Oracle top-1 recall & $-6.8$ & $[-8.9,-4.6]$ \\
\bottomrule
\end{tabular}
\normalsize
\caption{Paired scenario-level bootstrap intervals for proposal-only
IGD minus Base. The bootstrap uses 10,000 resamples and preserves
method and seed pairing. Recall differences are percentage points.}
\label{tab:proposal_only_bootstrap}
\end{table}

The published-score change is deliberately interpreted as a small
decrease rather than a large performance loss. The substantive result
is that a 1.03-point improvement in available-best quality is not
converted into published quality, causing regret to increase by
1.21 points and oracle top-1 recall to decrease by 6.8 percentage
points.

\subsection{Independent Counterfactual Scores by Seed}

Table~\ref{tab:counterfactual_seed_results} reports the published score,
oracle-best score, candidate-publication regret, and oracle top-1 recall
for all five independently trained seeds.

\begin{table*}[!t]
\centering
\small
\begin{tabular}{lcccc}
\toprule
Seed & Published $S_{\mathrm{cf}}$ & Oracle-best & $\Delta_{\mathrm{cf}}$ & Top-1 recall \\
\midrule
0 & 92.20 & 92.86 & 0.66 & 85.9 \\
1 & 92.46 & 93.03 & 0.57 & 87.2 \\
2 & 92.28 & 92.92 & 0.64 & 86.4 \\
3 & 92.37 & 92.98 & 0.61 & 86.9 \\
4 & 92.24 & 92.86 & 0.62 & 86.6 \\
\midrule
Mean & 92.31 & 92.93 & 0.62 & 86.6 \\
Std. & 0.10 & 0.07 & 0.03 & 0.49 \\
\bottomrule
\end{tabular}
\normalsize
\caption{Per-seed counterfactual candidate diagnostics over $\Omega_{\mathrm{sel}}$. Oracle-best is taken over all generated candidates, including candidates rejected by online hard validity. Published and oracle-best scores use identical candidate rollouts; all-invalid/MRM states are reported separately. Std. is computed across seeds.}
\label{tab:counterfactual_seed_results}
\end{table*}

\section{Proposal and Publication Ablations}
\label{app:core_ablations}
\begin{figure}[!t]
\centering
\includegraphics[width=\columnwidth]{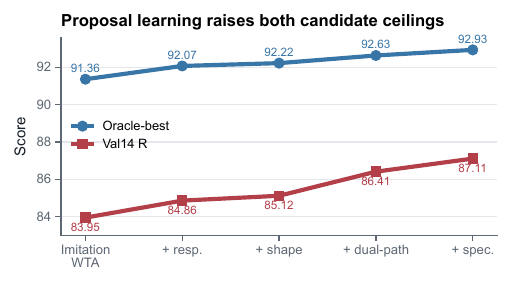}
\caption{Stage-controlled proposal learning with the full EGS selector
fixed. Execution-conditioned responsibility, differentiable geometry
shaping, and specialization progressively improve candidate availability
and reactive closed-loop performance. Values are five-seed means.}
\label{fig:appendix_proposal_progression}
\end{figure}

\begin{figure*}[!t]
\centering
\includegraphics[width=0.94\textwidth]{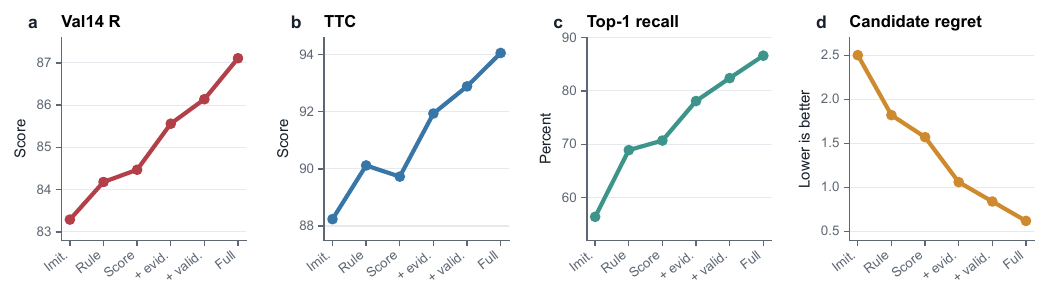}
\caption{Execution-grounded publication cascade with the full proposal generator
fixed. Learned scoring, online evidence, hard validity, and adaptive recurrent
execution progressively improve reactive and TTC scores while reducing
candidate-publication regret. Values are five-seed means.}
\label{fig:appendix_selection_cascade}
\end{figure*}

This section expands the stage-controlled ablations summarized in the main
paper. Table~\ref{tab:mechanism_ablation} separates proposal learning from
publication, Figures~\ref{fig:appendix_proposal_progression}
and~\ref{fig:appendix_selection_cascade} visualize the complete cascades,
and Table~\ref{tab:temporal_execution} isolates recurrent continuity.
The rule-cost row directly minimizes the online analytic evidence cost
in Eq.~\eqref{eq:app_online_rule_cost}, whereas the learned EGS rows
publish by maximizing the context-dependent utility $u_\phi$. The
comparison isolates the value of learned candidate--context interactions
beyond fixed-cost reranking.
Proposal-side rows hold the full EGS selector fixed, publication-side rows freeze the same
proposal generator, and temporal rows differ only in the continuity
mechanism.

\begin{table*}[!t]
\centering
\small
\setlength{\tabcolsep}{2.7pt}
\textbf{(a) Execution-conditioned diffusion proposal learning}
\resizebox{\textwidth}{!}{%
\begin{tabular}{lccccccccc}
\toprule
Proposal training & $\mathcal L_{\mathrm{resp}}$ & $\mathcal L_{\mathrm{shape}}$ & $\mathcal L_{\mathrm{spec}}$ & Oracle-best & Feasible ratio & Feasible APD & All-invalid & Val14 NR & Val14 R \\
\midrule
IGD + imitation WTA & -- & -- & -- & 91.36 & 86.60 & 1.13 & 2.17 & 90.84 & 83.95 \\
IGD + responsibility & \checkmark & -- & -- & 92.07 & 89.70 & 1.31 & 1.46 & 91.27 & 84.86 \\
IGD + geometry shaping & -- & \checkmark & -- & 92.22 & 90.60 & 1.39 & 1.18 & 91.43 & 85.12 \\
IGD + dual-path & \checkmark & \checkmark & -- & 92.63 & 92.10 & 1.52 & 0.82 & 92.05 & 86.41 \\
IGD + dual-path + specialization & \checkmark & \checkmark & \checkmark & 92.93 & \textbf{93.40} & \textbf{1.66} & \textbf{0.68} & \textbf{92.37} & \textbf{87.11} \\
\bottomrule
\end{tabular}%
}

\textbf{(b) Recurrent Execution-Grounded Publication}
\resizebox{\textwidth}{!}{%
\begin{tabular}{lccccccccc}
\toprule
Publication interface & Scorer & Evidence & Hard validity & Temporal & Val14 NR & Val14 R & TTC & Oracle top-1 recall & $\Delta_{\mathrm{cf}}$ \\
\midrule
Imitation selector & -- & -- & -- & None & 90.49 & 83.29 & 88.24 & 56.4 & 2.50 \\
Rule-cost selector & -- & \checkmark & \checkmark & None & 91.17 & 84.18 & 90.12 & 68.9 & 1.82 \\
Learned scorer only & \checkmark & -- & -- & None & 91.42 & 84.47 & 89.73 & 70.7 & 1.57 \\
Scorer + evidence & \checkmark & \checkmark & -- & None & 91.93 & 85.56 & 91.94 & 78.1 & 1.06 \\
+ hard validity & \checkmark & \checkmark & \checkmark & None & 92.15 & 86.14 & 92.89 & 82.4 & 0.84 \\
+ fixed temporal feedback & \checkmark & \checkmark & \checkmark & Fixed & 92.24 & 86.49 & 93.33 & 83.7 & 0.75 \\
Full EGS selector & \checkmark & \checkmark & \checkmark & Adaptive & \textbf{92.37} & \textbf{87.11} & \textbf{94.06} & 86.6 & \textbf{0.62} \\
\bottomrule
\end{tabular}%
}
\normalsize
\caption{Five-seed controlled ablation on Val14. Panel (a)
changes proposal learning while keeping the full EGS selector fixed. Panel
(b) freezes the same full proposal generator and changes only
the publication interface. Oracle quantities use the complete generated set, including candidates
rejected by online hard validity.}
\label{tab:mechanism_ablation}
\end{table*}

\begin{table*}[!t]
\centering
\small
\setlength{\tabcolsep}{2.8pt}
\begin{tabular}{@{}lccccccccc@{}}
\toprule
Temporal strategy & Val14 R & TTC & \shortstack{Plan dev.\\$\downarrow$} & \shortstack{Switch\\$\downarrow$} & \shortstack{Hazard\\delay $\downarrow$} & \shortstack{Hazard\\success $\uparrow$} & \shortstack{Avg.\\speed} & \shortstack{Unnec.\\stop $\downarrow$} & \shortstack{Route\\progress $\uparrow$} \\
\midrule
No temporal feedback & 86.14 & 92.89 & 0.313 & 13.6 & \textbf{0.16} & 91.9 & 6.42 & 3.42 & 89.82 \\
Fixed temporal feedback & 86.49 & 93.33 & 0.229 & 8.8 & 0.31 & 90.8 & 6.38 & 3.37 & 89.91 \\
Adaptive Continuity Gate & \textbf{87.11} & \textbf{94.06} & \textbf{0.205} & \textbf{7.3} & 0.18 & \textbf{93.4} & 6.39 & \textbf{3.19} & \textbf{90.05} \\
\bottomrule
\end{tabular}
\normalsize
\caption{Recurrent stability, hazard responsiveness, and
execution efficiency on Val14. Hazard delay is in seconds; rate
columns are percentages.}
\label{tab:temporal_execution}
\end{table*}

\section{Closed-Loop Consequences and Diagnostic Validity}
\label{app:counterfactual_validity}

This section tests the diagnostic conclusion through commit-horizon
sensitivity and forced-prefix interventions with identical resumed
replanning. Neither analysis changes training, checkpoint selection,
or online publication.

\subsection{Commit-Horizon Sensitivity with a Fixed Candidate Set}
\label{app:commit_horizon_sensitivity}

At each eligible diagnostic state, the frozen full proposal generator
produces one candidate set. The imitation selector, rule-cost
selector, and the full EGS selector operate on exactly this same set. For each
commit horizon
$H\in\{0.5,1.0,2.0,4.0,8.0\}$~s, every candidate is independently
committed from the same cloned simulator state for $H$ seconds under
matched conditions. Oracle-best is consequently identical across
selectors within each horizon. Absolute scores are compared across
selectors only within the same horizon because different horizons
evaluate different rollout durations.

\begin{table*}[!t]
\centering
\small
\setlength{\tabcolsep}{5pt}
\begin{tabular}{llcccc}
\toprule
Commit horizon &
Selector &
Published score $\uparrow$ &
Oracle-best $\uparrow$ &
Regret $\downarrow$ &
Top-1 recall (\%) $\uparrow$ \\
\midrule
0.5~s & Imitation selector    & 96.84 & 97.12 & 0.28 & 67.3 \\
0.5~s & Rule-cost selector    & 96.93 & 97.12 & 0.19 & 74.8 \\
0.5~s & Full EGS selector     & \textbf{97.03} & 97.12 & \textbf{0.09} & \textbf{88.1} \\
\midrule
1.0~s & Imitation selector    & 95.38 & 95.94 & 0.56 & 62.5 \\
1.0~s & Rule-cost selector    & 95.56 & 95.94 & 0.38 & 72.1 \\
1.0~s & Full EGS selector     & \textbf{95.77} & 95.94 & \textbf{0.17} & \textbf{87.4} \\
\midrule
2.0~s & Imitation selector    & 93.71 & 94.78 & 1.07 & 59.1 \\
2.0~s & Rule-cost selector    & 94.03 & 94.78 & 0.75 & 70.6 \\
2.0~s & Full EGS selector     & \textbf{94.43} & 94.78 & \textbf{0.35} & \textbf{86.9} \\
\midrule
4.0~s & Imitation selector    & 91.52 & 93.18 & 1.66 & 56.8 \\
4.0~s & Rule-cost selector    & 92.03 & 93.18 & 1.15 & 69.2 \\
4.0~s & Full EGS selector     & \textbf{92.63} & 93.18 & \textbf{0.55} & \textbf{86.1} \\
\midrule
8.0~s & Imitation selector    & 90.43 & 92.93 & 2.50 & 56.4 \\
8.0~s & Rule-cost selector    & 91.11 & 92.93 & 1.82 & 68.9 \\
8.0~s & Full EGS selector     & \textbf{92.31} & 92.93 & \textbf{0.62} & \textbf{86.6} \\
\bottomrule
\end{tabular}
\normalsize
\caption{Commit-horizon sensitivity under a fixed candidate set.
All selectors operate on the same candidates, so oracle-best is
identical across selectors within each horizon. Scores should not be
compared directly across horizons because the rollout duration
changes. Values follow the same five-seed protocol as the primary
evaluation.}
\label{tab:commit_horizon_sensitivity}
\end{table*}

The full EGS selector achieves the highest published score and lowest regret at
every horizon. Relative to the imitation selector, its regret
reductions are 67.9\%, 69.6\%, 67.3\%, 66.9\%, and 75.2\% from
0.5 to 8.0~s. Its top-1 recall remains between 86.1\% and 88.1\%,
whereas imitation recall decreases from 67.3\% to 56.4\% as
longer-term execution consequences become observable. The publication
advantage is therefore not created by the original 8.0-s diagnostic
horizon.

\subsection{Prefix Intervention with Resumed Replanning}
\label{app:prefix_resumed_replanning}

For each diagnostic state, we clone the simulator state, ego state,
surrounding-agent states, route, map context, random state, and planner
recurrent state. We first force the ego vehicle to execute the prefix
of one candidate for
$H\in\{0.5,1.0,2.0,4.0\}$~s. Every branch then resumes normal
10-Hz replanning with the same frozen full \ours planner. Branch
comparisons at each horizon use matched eligible states.

Eligibility is determined independently of branch performance.
Collision and task termination remain evaluation outcomes and use a
fixed terminal-state score for the remainder of the 8.0-s window.
States are excluded only when the cloned state or fields required to
initialize and evaluate all compared branches are unavailable.
Consequently, all prefix sources share the same matched eligible set
within each horizon. Scenario counts denote unique scenario tokens.

The total branch-evaluation horizon is fixed to 8.0~s from the
original diagnostic state. A 0.5-s forced prefix is followed by
7.5~s of resumed replanning; 1.0-, 2.0-, and 4.0-s prefixes are
followed by 7.0, 6.0, and 4.0~s, respectively. Hence, every prefix
source and prefix horizon is evaluated over the same total duration.

The frozen full planner comprises the same trained checkpoint, IGD
proposal generator, EGS selector, forecast head, allocation head,
hard-validity thresholds, continuity gate, and MRM policy.
After the forced prefix, no branch retains its original selector; all
branches use the identical full \ours planner. The only intervention is
the initially forced prefix, after which the physical scene evolves
from the state induced by its execution.

\paragraph{Recurrent-state initialization.}
The forced prefix is treated as the previously published trajectory at
resumption. Its unexecuted suffix, when available, is aligned to the
current ego frame, resampled onto the planning grid, and supplied as
previous-publication context. The ego state, surrounding-agent states,
simulator random state, and elapsed time are advanced by the
forced-prefix rollout. Resumed planning therefore starts from the
physical and recurrent state induced by the executed prefix rather
than from the original diagnostic state.

\paragraph{Retrospective oracle prefix.}
The oracle prefix is selected from the same candidate set generated
at the original diagnostic state. For every candidate, we evaluate
the branch obtained by forcing its $H$-second prefix and resuming the
frozen full \ours planner for the remaining $8.0-H$ seconds. The
oracle-best prefix is the candidate with the highest total
resumed-branch score over the complete 8.0-s window. This oracle is
used only for diagnosis and is unavailable to training and online
inference.

\paragraph{Degradation and recovery.}
Route-progress deficit is
$\max(0,p_{\mathrm{ref}}-p_{\mathrm{branch}})$, where
$p_{\mathrm{ref}}$ is the fixed selector-independent reference used
by the evaluator over the same elapsed window. A branch is degraded
if, during the forced prefix or the first 1.0~s after it, any of the
following occurs: TTC below 2.0~s, route-progress deficit of at least
3.0~m, selected-trajectory risk evidence of at least 0.50, no
hard-valid candidate, MRM activation, or collision. For each horizon,
the common degraded set contains a diagnostic state when at least one
compared branch is degraded.

Every prefix source is evaluated over this same common denominator. A
branch is recovered only if, throughout the final 3.0~s of the 8.0-s
window, it has no collision or MRM activation, TTC of at least 2.0~s,
at least one hard-valid candidate at every replanning cycle, risk
evidence below 0.50, and route-progress deficit below 3.0~m. These
thresholds are diagnostic and do not modify online hard validity or
publication.

\begin{table*}[!t]
\centering
\small
\setlength{\tabcolsep}{2.2pt}
\begin{tabular}{@{}llrrrrrr@{}}
\toprule
Prefix &
Initial source &
Scenarios &
\shortstack{Eligible\\states} &
Score $\uparrow$ &
\shortstack{TTC\\score $\uparrow$} &
\shortstack{Collision\\(\%) $\downarrow$} &
\shortstack{Progress\\score $\uparrow$} \\
\midrule
0.5~s & Imitation prefix & 1,118 & 111,820 & 86.44 & 92.81 & 0.31 & 89.61 \\
0.5~s & Rule-cost prefix & 1,118 & 111,820 & 86.58 & 93.02 & 0.27 & 89.72 \\
0.5~s & EGS prefix       & 1,118 & 111,820 & \textbf{86.95} & \textbf{93.41} & \textbf{0.19} & \textbf{89.91} \\
0.5~s & Oracle prefix    & 1,118 & 111,820 & 87.12 & 93.56 & 0.17 & 90.02 \\
\midrule
1.0~s & Imitation prefix & 1,118 & 111,406 & 86.05 & 92.38 & 0.39 & 89.34 \\
1.0~s & Rule-cost prefix & 1,118 & 111,406 & 86.34 & 92.76 & 0.32 & 89.55 \\
1.0~s & EGS prefix       & 1,118 & 111,406 & \textbf{86.96} & \textbf{93.57} & \textbf{0.18} & \textbf{89.96} \\
1.0~s & Oracle prefix    & 1,118 & 111,406 & 87.20 & 93.76 & 0.15 & 90.09 \\
\midrule
2.0~s & Imitation prefix & 1,117 & 110,736 & 85.41 & 91.74 & 0.51 & 88.92 \\
2.0~s & Rule-cost prefix & 1,117 & 110,736 & 85.88 & 92.24 & 0.42 & 89.21 \\
2.0~s & EGS prefix       & 1,117 & 110,736 & \textbf{86.83} & \textbf{93.44} & \textbf{0.21} & \textbf{89.88} \\
2.0~s & Oracle prefix    & 1,117 & 110,736 & 87.16 & 93.82 & 0.16 & 90.13 \\
\midrule
4.0~s & Imitation prefix & 1,115 & 109,284 & 84.62 & 90.83 & 0.68 & 88.21 \\
4.0~s & Rule-cost prefix & 1,115 & 109,284 & 85.27 & 91.56 & 0.55 & 88.73 \\
4.0~s & EGS prefix       & 1,115 & 109,284 & \textbf{86.52} & \textbf{93.10} & \textbf{0.26} & \textbf{89.67} \\
4.0~s & Oracle prefix    & 1,115 & 109,284 & 86.98 & 93.61 & 0.18 & 90.01 \\
\bottomrule
\end{tabular}
\normalsize
\caption{Closed-loop branch evaluation after prefix execution and
resumed replanning. Each branch starts from the same cloned state,
executes the indicated prefix, and then uses the same frozen full
\ours planner for the remainder of the fixed 8.0-s window. Counts are
unique scenario tokens and matched eligible diagnostic states. Bold
marks the strongest online prefix source; the retrospective oracle is
shown only as a diagnostic upper bound.}
\label{tab:prefix_resumed_replanning}
\end{table*}

\begin{table}[!t]
\centering
\small
\setlength{\tabcolsep}{2.4pt}
\begin{tabular}{@{}lccccc@{}}
\toprule
Prefix &
Common degraded &
Imitation &
Rule cost &
EGS &
Oracle \\
\midrule
0.5~s & 18,734 & 84.7 & 85.5 & \textbf{87.8} & 88.4 \\
1.0~s & 21,968 & 82.9 & 84.1 & \textbf{88.3} & 89.0 \\
2.0~s & 27,615 & 79.8 & 81.6 & \textbf{87.1} & 88.7 \\
4.0~s & 35,902 & 75.4 & 78.3 & \textbf{85.9} & 87.6 \\
\bottomrule
\end{tabular}
\normalsize
\caption{Recovery rates (\%) after resumed replanning. Within each
horizon, all prefix sources are evaluated over the same common
degraded-state denominator. Bold marks the strongest online prefix
source; the oracle remains retrospective.}
\label{tab:prefix_recovery}
\end{table}

\begin{table}[!t]
\centering
\small
\setlength{\tabcolsep}{4.2pt}
\begin{tabular}{llrrrr}
\toprule
Seed & Prefix & Imitation & Rule cost & EGS & Oracle \\
\midrule
0 & 0.5~s & 86.34 & 86.51 & 86.86 & 87.04 \\
1 & 0.5~s & 86.55 & 86.71 & 87.08 & 87.22 \\
2 & 0.5~s & 86.39 & 86.56 & 86.88 & 87.06 \\
3 & 0.5~s & 86.47 & 86.64 & 86.99 & 87.15 \\
4 & 0.5~s & 86.45 & 86.49 & 86.95 & 87.12 \\
\midrule
0 & 1.0~s & 85.97 & 86.27 & 86.89 & 87.14 \\
1 & 1.0~s & 86.24 & 86.50 & 87.13 & 87.36 \\
2 & 1.0~s & 86.02 & 86.31 & 86.90 & 87.18 \\
3 & 1.0~s & 86.11 & 86.40 & 87.02 & 87.26 \\
4 & 1.0~s & 85.91 & 86.23 & 86.86 & 87.06 \\
\midrule
0 & 2.0~s & 85.32 & 85.79 & 86.75 & 87.08 \\
1 & 2.0~s & 85.58 & 86.03 & 87.05 & 87.36 \\
2 & 2.0~s & 85.37 & 85.85 & 86.78 & 87.12 \\
3 & 2.0~s & 85.49 & 85.96 & 86.91 & 87.25 \\
4 & 2.0~s & 85.29 & 85.77 & 86.65 & 86.99 \\
\midrule
0 & 4.0~s & 84.55 & 85.19 & 86.44 & 86.90 \\
1 & 4.0~s & 84.83 & 85.50 & 86.75 & 87.22 \\
2 & 4.0~s & 84.61 & 85.24 & 86.48 & 86.93 \\
3 & 4.0~s & 84.73 & 85.39 & 86.61 & 87.06 \\
4 & 4.0~s & 84.38 & 85.03 & 86.32 & 86.79 \\
\bottomrule
\end{tabular}
\normalsize
\caption{Seed-level resumed scores for each prefix horizon and source.
Each row is an independently trained model evaluated under the matched
branch protocol.}
\label{tab:prefix_seed_scores}
\end{table}

\begin{table}[!t]
\centering
\small
\setlength{\tabcolsep}{4.0pt}
\begin{tabular}{llrr}
\toprule
Prefix & Comparison & Mean diff. & Paired 95\% CI \\
\midrule
0.5~s & EGS $-$ imitation & $+0.51$ & $[+0.33,+0.69]$ \\
       & EGS $-$ rule cost & $+0.37$ & $[+0.21,+0.53]$ \\
       & Oracle $-$ EGS    & $+0.17$ & $[+0.06,+0.29]$ \\
\midrule
1.0~s & EGS $-$ imitation & $+0.91$ & $[+0.66,+1.15]$ \\
       & EGS $-$ rule cost & $+0.62$ & $[+0.39,+0.84]$ \\
       & Oracle $-$ EGS    & $+0.24$ & $[+0.10,+0.39]$ \\
\midrule
2.0~s & EGS $-$ imitation & $+1.42$ & $[+1.08,+1.75]$ \\
       & EGS $-$ rule cost & $+0.95$ & $[+0.66,+1.23]$ \\
       & Oracle $-$ EGS    & $+0.33$ & $[+0.15,+0.52]$ \\
\midrule
4.0~s & EGS $-$ imitation & $+1.90$ & $[+1.44,+2.34]$ \\
       & EGS $-$ rule cost & $+1.25$ & $[+0.87,+1.62]$ \\
       & Oracle $-$ EGS    & $+0.46$ & $[+0.22,+0.70]$ \\
\bottomrule
\end{tabular}
\normalsize
\caption{Scenario-clustered paired bootstrap intervals for resumed
score. Scenario tokens are resampled as clusters while retaining their
associated method, seed, and diagnostic-state outcomes.}
\label{tab:prefix_bootstrap}
\end{table}

Relative to the imitation prefix, the EGS-minus-imitation
resumed-score differences are 0.51, 0.91, 1.42, and 1.90 points for
horizons of 0.5, 1.0, 2.0, and 4.0~s. The corresponding TTC
differences are 0.60, 1.19, 1.70, and 2.27 points; collision
differences are $-0.12$, $-0.21$, $-0.30$, and $-0.42$ percentage
points; and route-progress differences are 0.30, 0.62, 0.96, and
1.46 points. The full EGS selector is 0.17, 0.24, 0.33, and 0.46 points below the
retrospective oracle. All paired resumed-score intervals exclude zero
(Table~\ref{tab:prefix_bootstrap}). Recovery uses the common
degraded-state denominator within each horizon
(Table~\ref{tab:prefix_recovery}).

\subsection{Scenario-Wise Effects of the Initial Prefix}
\label{app:scenario_prefix_effects}

\begin{figure}[!t]
\centering
\includegraphics[width=\columnwidth]{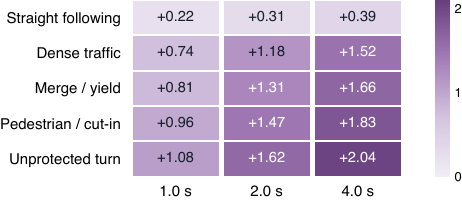}
\caption{Scenario-wise resumed-replanning gains. Each cell reports the
five-seed EGS-minus-imitation score difference. Columns show 1.0-, 2.0-,
and 4.0-s forced-prefix horizons; values should be compared within a
column because longer forced prefixes leave less time for the common
planner to recover.}
\label{fig:supp_scenario_gain_heatmap}
\end{figure}

Figure~\ref{fig:supp_scenario_gain_heatmap} localizes the downstream
consequence of the initial publication. The branch gain is only
0.22--0.39 points in straight following but increases to 1.08--2.04
points in unprotected turns. Dense traffic, merge or yield, and
pedestrian or cut-in scenes show the same pattern. The effect is
therefore concentrated where the initially executed trajectory changes
the interaction state and the difficulty of subsequent replanning.

\section{Reliability, Failure Boundaries, and Cost}
\label{sec:supp_reliability_generalization}
\label{sec:app_safety}

This section reports reliability, forecast-dependent failure boundaries,
scenario dependence, and compute controls supporting the main paper.

\subsection{Does Recurrent Publication Remain Stable Under Forecast Failure?}
\label{sec:execution_reliability}

\begin{figure*}[!t]
\centering
\includegraphics[width=0.93\textwidth]{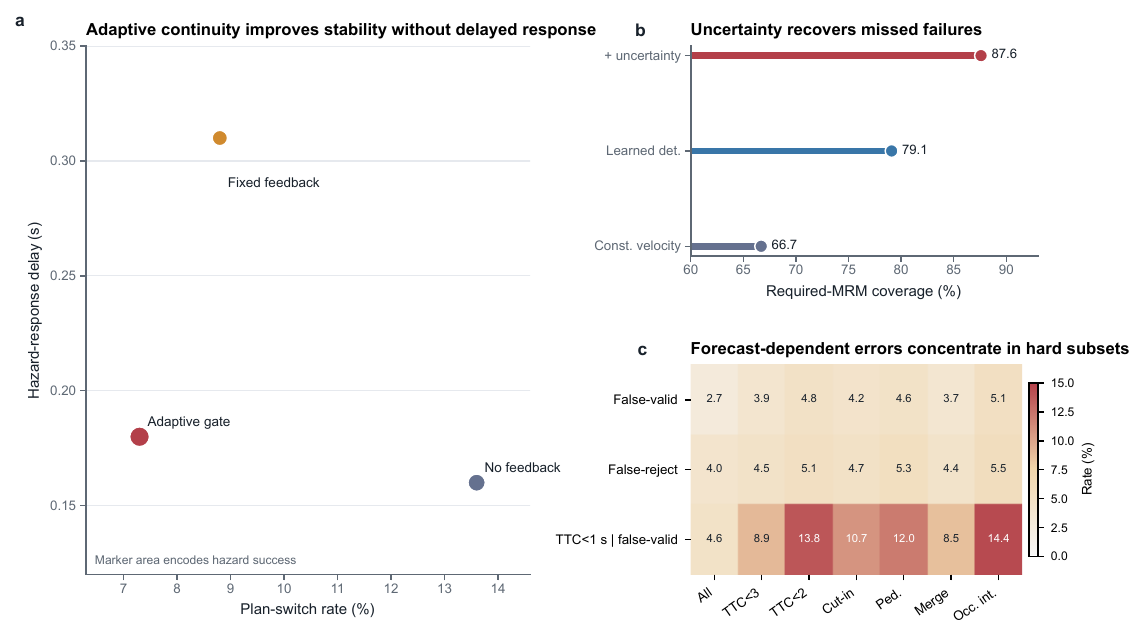}
\caption{Stability, fallback coverage, and forecast-dependent failure. (a) The Adaptive Continuity Gate reduces plan switching without the hazard-response delay introduced by fixed temporal feedback; marker area encodes hazard success. (b) Learned forecasting with uncertainty increases coverage of counterfactually required MRM events. (c) False-valid, false-reject, and severe low-TTC rates among false-valid decisions increase in high-risk subsets, especially occluded intersections. Rates and scores are pooled or averaged according to the units defined in Appendix Sec.~\ref{app:statistics}.}
\label{fig:execution_reliability}
\end{figure*}

Figure~\ref{fig:execution_reliability}(a) evaluates the
stability--responsiveness trade-off. Fixed temporal feedback reduces
plan switching from 13.6\% to 8.8\% but increases hazard-response delay
from 0.16 to 0.31~s. The Adaptive Continuity Gate further lowers
switching to 7.3\%, limits delay to 0.18~s, and raises hazard success to
93.4\%, showing that its gain does not come from blindly retaining the
previous publication.

The fallback analysis distinguishes detected proposal failure from
guaranteed safety. Uncertainty-aware forecasting covers 87.6\% of
required-MRM events, with 0.09\% missed and 0.02\% false activations.
Errors nevertheless increase in harder subsets: false-valid decisions
reach 5.14\% in occluded intersections, and 14.4\% of them have
counterfactual TTC below one second. MRM therefore responds
deterministically to detected failure, but forecast errors can still
prevent correct activation.

\subsection{Where Do the Gains Occur, and What Do They Cost?}
\label{sec:generalization_efficiency}

\begin{figure*}[!t]
\centering
\includegraphics[width=0.93\textwidth]{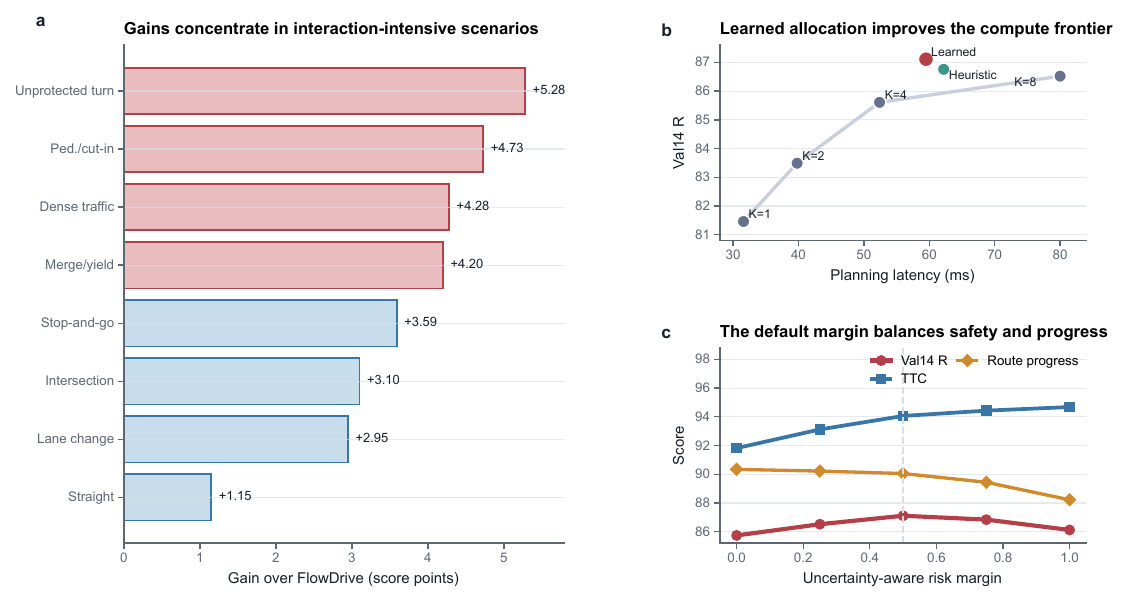}
\caption{Scenario dependence, computation, and parameter boundary. (a) Scenario-wise gains over FlowDrive are largest in unprotected turns, pedestrian or cut-in events, dense traffic, and merge or yield interactions. (b) Learned candidate allocation improves the Val14 reactive-score versus latency frontier relative to fixed budgets and heuristic adaptation. (c) Increasing the uncertainty-aware risk margin improves TTC, but excessive margins reduce route progress and overall reactive performance; 0.50 is the fixed default. Values are five-seed means.}
\label{fig:generalization_efficiency}
\end{figure*}

The scenario analysis in Fig.~\ref{fig:generalization_efficiency}(a) localizes the improvement. The gain over FlowDrive is 1.15 points in straight following but rises to 4.20 in merge or yield scenes, 4.28 in dense traffic, 4.73 for pedestrian or cut-in events, and 5.28 in unprotected turns. This concentration supports the proposed mechanism because interaction changes the value of the trajectory that is actually published.

The gain has a measurable computational cost. \ours adds 5.2 million parameters and runs in 59.5~ms on one NVIDIA GeForce RTX~4090D, compared with 50.7~ms for Diffusion Planner and 47.9~ms for FlowDrive. Learned allocation nevertheless dominates fixed $K=8$: it reaches 87.11 Val14 R at 59.5~ms, whereas fixed $K=8$ reaches 86.52 at 80.0~ms. At the matched average decoded budget $\bar K=5.5$, joint learned budgeting and priority reaches 92.37/87.11 NR/R, compared with 92.11/86.77 for the strongest disentangled control.

Risk-margin sensitivity identifies a bounded operating region. Raising the margin from 0 to 0.50 improves Val14 R from 85.74 to 87.11 and TTC from 91.82 to 94.06. Larger margins continue to improve TTC, but reduce route progress, comfort, and aggregate closed-loop performance. The safety gain is therefore not monotonic and cannot be attributed to unbounded conservatism. The execution-event analysis below shows that the full system maintains average speed while reducing unnecessary stops, high jerk, and emergency braking.

\paragraph{Mechanism and matched-compute controls.}
These controls isolate forecast quality, adaptive compute, scene-conditioned
query priority, and latent-query specialization. Matched-compute rows share the
same average candidate count, decoder, hardware, and evaluation scenarios.

\subsection{Online Forecast and Validity}

\begin{table*}[!t]
\centering
\small
\setlength{\tabcolsep}{2.2pt}
\begin{tabular}{lcccccc}
\toprule
Forecast & minADE & minFDE & Val14 R & TTC & False-valid & False-reject \\
\midrule
Constant velocity & 1.10 & 2.36 & 86.02 & 92.76 & 4.80 & 3.20 \\
Learned deterministic & 0.82 & 1.72 & 86.66 & 93.51 & 3.50 & 3.40 \\
Learned + uncertainty & 0.82 & 1.72 & \textbf{87.11} & \textbf{94.06} & \textbf{2.70} & 4.00 \\
\bottomrule
\end{tabular}
\normalsize
\caption{Forecast accuracy is measured in meters over the 8.0-s horizon on interaction-relevant agents. Validity rates are percentages.}
\label{tab:forecast_ablation}
\end{table*}

\begin{figure*}[!t]
\centering
\includegraphics[width=0.98\textwidth]{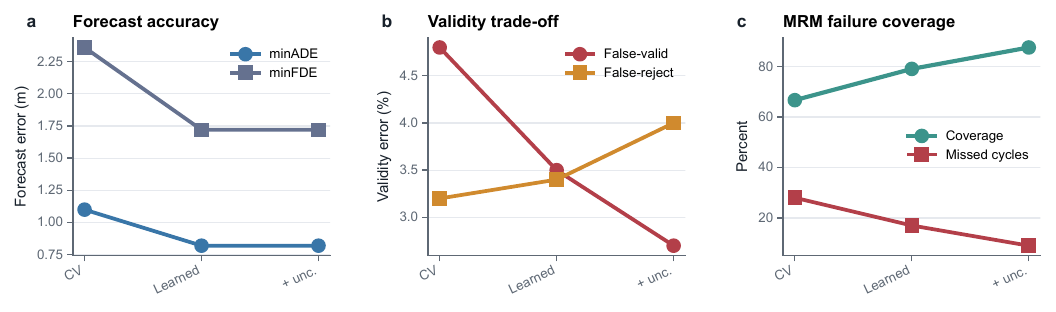}
\caption{Forecast and validity diagnostics. (a) Learned forecasting reduces minADE and minFDE relative to constant velocity. (b) Uncertainty reduces false-valid decisions at the cost of a moderate increase in false rejection. (c) The same uncertainty-aware interface raises required-MRM coverage and reduces missed-MRM cycles. Exact values appear in Tables~\ref{tab:forecast_ablation} and~\ref{tab:mrm_coverage}.}
\label{fig:appendix_forecast_validity}
\end{figure*}

\subsection{Matched-Compute Budget and Candidate Priority}

A random matched budget samples $K=2,4,8$ with probabilities
$0.125,0.4375,0.4375$, respectively, so
$\mathbb E[K]=5.5$. All rows use the same 25-step decoder,
hardware, and five seeds. Fixed order uses query index; random
order is redrawn independently at every planning state.
Scene-conditioned priority is predicted from $(h_t,c_t,q_m)$
before diffusion decoding, so only the selected query prefix incurs
decoder cost.

\begin{table*}[!t]
\centering
\small
\setlength{\tabcolsep}{3.2pt}
\resizebox{\textwidth}{!}{%
\begin{tabular}{lcccccccc}
\toprule
Budget policy & Candidate order & Avg. $K$ & Latency & Budget regret & Oracle-best & Feasible ratio & Val14 NR & Val14 R \\
\midrule
Random matched & Random & 5.5 & 59.4 & 0.43 & 92.38 & 90.81 & 91.76 & 86.19 \\
Random matched & Scene-conditioned priority & 5.5 & 59.1 & 0.27 & 92.61 & 91.75 & 92.04 & 86.48 \\
Learned budget & Fixed query order & 5.5 & 59.6 & 0.22 & 92.69 & 92.26 & 92.11 & 86.77 \\
Learned budget & Random order & 5.5 & 59.3 & 0.30 & 92.55 & 91.94 & 91.97 & 86.59 \\
Joint learned allocation & Scene-conditioned priority & 5.5 & 59.5 & \textbf{0.12} & 92.93 & \textbf{93.40} & \textbf{92.37} & \textbf{87.11} \\
\bottomrule
\end{tabular}%
}
\normalsize
\caption{Matched-compute controls disentangle budget prediction, candidate
priority, and nested ordering. Every row has $\bar K=5.5$ and uses the same
25-step decoder, hardware, and five seeds.}
\label{tab:budget_matched_compute}
\end{table*}

Both factors contribute: scene-conditioned priority improves the random
budget by 0.28/0.29 NR/R, while learned budgeting improves the
fixed-order control. Their joint use provides a further
0.26/0.34 gain over the strongest disentangled control at
nearly identical latency.

\subsection{Latent-Intent Specialization}

\begin{table*}[!t]
\centering
\small
\setlength{\tabcolsep}{2.2pt}
\begin{tabular}{lcccccc}
\toprule
Variant & Entropy & Slots & Intent cons. & Feasible & Oracle & Collapse \\
\midrule
Imitation WTA & 0.73 & 4.58 & 87.94 & 86.60 & 91.36 & 18.2 \\
+ responsibility & 0.82 & 5.47 & 90.21 & 89.70 & 92.07 & 12.9 \\
+ usage balance & 0.88 & 6.19 & 92.16 & 91.84 & 92.44 & 8.4 \\
Full specialization & 0.91 & 6.63 & 94.20 & 93.40 & 92.93 & 5.8 \\
\bottomrule
\end{tabular}
\normalsize
\caption{Latent-query specialization diagnostics.}
\label{tab:intent_specialization}
\end{table*}

\begin{figure*}[!t]
\centering
\includegraphics[width=0.98\textwidth]{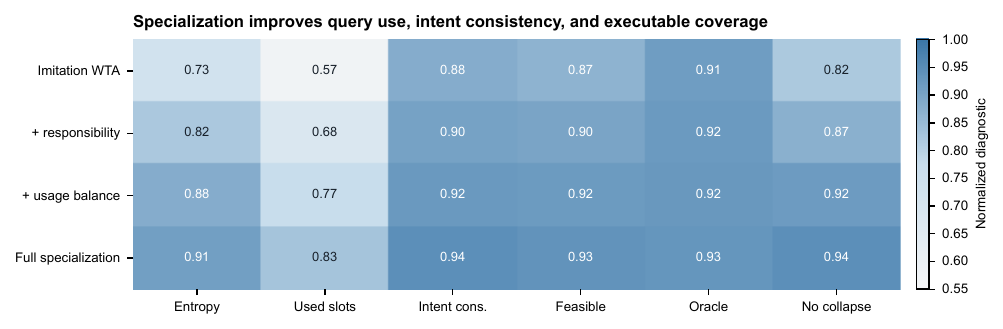}
\caption{Normalized latent-intent diagnostics. Used slots are divided by eight; percentage metrics are divided by 100; ``no collapse'' is one minus the collapse rate. The heatmap complements the exact unnormalized values in Table~\ref{tab:intent_specialization}.}
\label{fig:appendix_intent_specialization}
\end{figure*}

\paragraph{Fallback and forecast-dependent failure.}
We separately evaluate proposal failure, deterministic fallback benefit,
and forecast-dependent trigger coverage. This separation prevents MRM from
being interpreted as formal verification or a guaranteed safety mechanism.

\subsection{Candidate Budget and Runtime}

\begin{table*}[!t]
\centering
\small
\resizebox{\textwidth}{!}{%
\begin{tabular}{lcccccccc}
\toprule
Policy & Avg. $K$ & Raw APD & Feasible APD & Feasible ratio & Total ms & Failure/MRM & Val14 NR & Val14 R \\
\midrule
$K=1$ & 1.0 & 0.00 & 0.00 & 78.40 & 31.6 & 2.84 & 88.96 & 81.46 \\
$K=2$ & 2.0 & 0.83 & 0.63 & 84.00 & 39.8 & 2.02 & 90.18 & 83.49 \\
$K=4$ & 4.0 & 1.42 & 1.15 & 89.20 & 52.4 & 1.31 & 91.25 & 85.61 \\
$K=8$ & 8.0 & \textbf{2.07} & 1.49 & 88.60 & 80.0 & 0.76 & 91.86 & 86.52 \\
Heuristic adaptive & 5.8 & 1.94 & 1.57 & 91.60 & 62.2 & 0.74 & 92.05 & 86.76 \\
Learned adaptive & 5.5 & 1.89 & \textbf{1.66} & \textbf{93.40} & 59.5 & \textbf{0.68} & \textbf{92.37} & \textbf{87.11} \\
\bottomrule
\end{tabular}%
}
\normalsize
\caption{Candidate budget, executable diversity, runtime, and complete proposal failure.}
\label{tab:budget_runtime}
\end{table*}

Figure~\ref{fig:generalization_efficiency}(b) summarizes the same score--latency frontier; Table~\ref{tab:budget_runtime} reports the complete numerical operating points.

\subsection{MRM Benefit and Forecast-Failure Coverage}

Enabling the deterministic MRM increases collision, TTC, progress, comfort,
and Val14 reactive scores from 99.73, 93.62, 99.82, 90.97, and 86.84 to
100.00, 94.06, 100.00, 90.99, and 87.11, respectively. This comparison
isolates the closed-loop benefit of fallback activation without interpreting
it as formal safety verification.

\begin{table*}[!t]
\centering
\small
\setlength{\tabcolsep}{2.6pt}
\begin{tabular}{@{}lccccccc@{}}
\toprule
Online forecast & \shortstack{Oracle-required\\MRM} & \shortstack{Online\\MRM} & Coverage & \shortstack{Missed\\MRM} & \shortstack{False\\MRM} & \shortstack{Collision\\among missed} & \shortstack{TTC $<1$ s\\among missed} \\
\midrule
Constant velocity & 0.84 & 0.61 & 66.7 & 0.28 & 0.05 & 8.1 & 27.4 \\
Learned deterministic & 0.79 & 0.65 & 79.1 & 0.17 & 0.03 & 3.8 & 16.2 \\
Learned + uncertainty & 0.76 & 0.68 & 87.6 & 0.09 & 0.02 & 0.0 & 9.3 \\
\bottomrule
\end{tabular}
\normalsize
\caption{Forecast-failure-aware MRM analysis. The first, second,
fifth, and sixth columns except coverage are percentages of all
planning cycles; the final two columns are percentages within
missed-MRM events. Coverage is the fraction of counterfactually
required MRM events detected online.}
\label{tab:mrm_coverage}
\end{table*}

The uncertainty-aware forecast improves MRM coverage while
keeping false activation low. MRM is therefore a deterministic
response to detected complete proposal failure, not a guarantee:
forecast misses can still prevent activation.

\subsection{High-Risk False-Valid Analysis}

\begin{table*}[!t]
\centering
\small
\setlength{\tabcolsep}{3.3pt}
\begin{tabular}{lccccc}
\toprule
Subset & Events & False-valid & False-reject & Collision among false-valid & TTC $<1$ s among false-valid \\
\midrule
All diagnostic states & 112,840 & 2.70 & 4.00 & 0.8 & 4.6 \\
Counterfactual TTC $<3$ s & 6,387 & 3.91 & 4.54 & 1.7 & 8.9 \\
Counterfactual TTC $<2$ s & 2,264 & 4.76 & 5.11 & 2.5 & 13.8 \\
Cut-in & 1,193 & 4.24 & 4.69 & 2.0 & 10.7 \\
Pedestrian conflict & 742 & 4.58 & 5.29 & 2.3 & 12.0 \\
Merge/yield & 1,527 & 3.67 & 4.38 & 1.2 & 8.5 \\
Occluded intersection & 906 & 5.14 & 5.55 & 2.8 & 14.4 \\
\bottomrule
\end{tabular}
\normalsize
\caption{Validity errors stratified by counterfactual risk and scenario type, pooled across five seeds. Errors rise in the hardest subsets, motivating uncertainty inflation and explicit reporting rather than a global average alone.}
\label{tab:high_risk_validity}
\end{table*}

\paragraph{Safety and scenario analyses.}
These analyses test whether the improvement is conservative,
parameter-sensitive, or limited to simple scenarios. They complement the
aggregate score with interpretable safety and interaction diagnostics.

\subsection{Official Val14 Sub-Metrics}
Across five independently trained models, \ours obtains an official Val14
score of 92.37, with TTC 94.06, collision 100.00, comfort 90.99, progress
100.00, drivable-area compliance 100.00, driving-direction compliance 99.57,
speed-limit compliance 100.00, and route progress 90.05.

\subsection{Non-Conservative Execution}
\begin{figure*}[!t]
\centering
\includegraphics[width=0.94\textwidth]{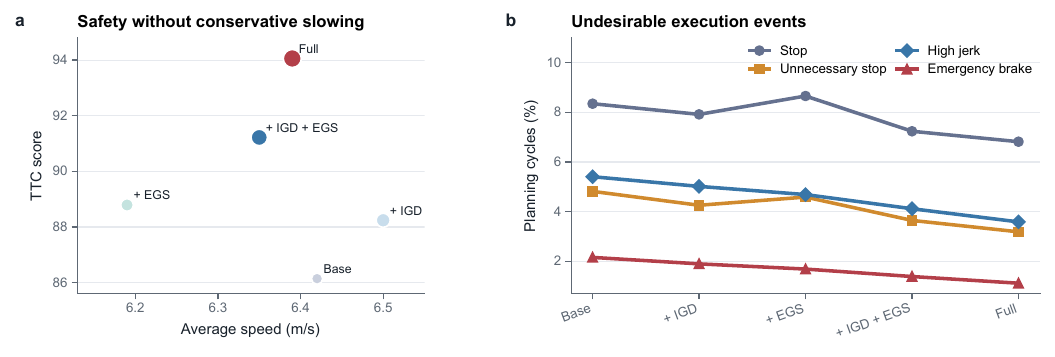}
\caption{Non-conservative execution diagnostics on Val14. (a) TTC improves
without reducing average speed; marker size encodes fewer unnecessary stops.
(b) The full model reduces stopping, unnecessary stopping, high-jerk events,
and emergency braking across planning cycles. Values are five-seed means.}
\label{fig:appendix_nonconservative}
\end{figure*}

Figure~\ref{fig:appendix_nonconservative} tests whether the safety gain is
obtained by slowing or stopping. The full model maintains an average speed of
6.39~m/s, close to 6.42~m/s for Base Diffusion, while increasing TTC from
86.14 to 94.06 and route progress from 87.80 to 90.05. It also reduces
unnecessary stops, high-jerk events, and emergency braking from
4.82/5.41/2.16\% to 3.19/3.59/1.12\%.

\subsection{Risk-Margin Sensitivity}
Figure~\ref{fig:generalization_efficiency}(c) shows that the default margin of
0.50 maximizes NR, R, and comfort jointly. Larger margins continue to improve
TTC but reduce route progress and overall closed-loop performance, exposing
the intended safety--efficiency trade-off rather than a monotonic tuning gain.

\subsection{Scenario-Wise Performance}
Figure~\ref{fig:generalization_efficiency}(a) localizes the gains over FlowDrive.
The largest improvements occur in unprotected turns, pedestrian/cut-in
events, dense traffic, and merge/yield scenarios, whereas straight following
shows the smallest gain. This concentration is consistent with the claim that
execution-grounded publication matters most when interaction changes the value
of the published trajectory.

\section{NAVSIM Additional Evidence}
\label{sec:app_navsim}
\label{app:navsim_protocol}

This section supports the cross-benchmark result in the main paper with
the exact training protocol, five independent runs, and consecutive-frame
qualitative evidence.

\subsection{Protocol and Seed-Level Results}
\label{sec:supp_navsim_protocol}

We use the official NAVSIM v1.1 release and train \ours from scratch
on \texttt{navtrain} without external data or pretrained checkpoints.
The fixed train/validation partition is log-disjoint and remains
unchanged across runs. For each independent run, we select the
checkpoint with the lowest validation total loss, computed using the
same objective as training. No \texttt{navtest} result is used for
checkpoint selection or hyperparameter tuning. We evaluate the
selected checkpoints on the official \texttt{navtest} split without
modifying the PDMS evaluator, LQR controller, metric cache, trajectory
sampling, or aggregation settings.

Table~\ref{tab:navsim_seed_results} reports five independently trained
models with seeds 0--4. All runs use the same data partition, training
budget, hyperparameters, and checkpoint-selection rule. The mean row
matches the NAVSIM result in the main paper. Standard deviations are
sample standard deviations across the five runs.

\begin{table}[!t]
\centering
\small
\setlength{\tabcolsep}{2.8pt}
\begin{tabular}{@{}lcccccc@{}}
\toprule
Seed & NC $\uparrow$ & DAC $\uparrow$ & TTC $\uparrow$ &
Comf. $\uparrow$ & EP $\uparrow$ & PDMS $\uparrow$ \\
\midrule
0 & 99.1 & 98.3 & 97.5 & 100.0 & 87.6 & 92.7 \\
1 & 99.0 & 97.9 & 97.1 & 100.0 & 87.4 & 92.2 \\
2 & 98.9 & 98.3 & 97.2 & 100.0 & 87.3 & 92.4 \\
3 & 99.1 & 98.3 & 97.5 & 100.0 & 87.7 & 92.8 \\
4 & 99.2 & 98.2 & 97.7 & 100.0 & 87.8 & 92.9 \\
\midrule
Mean & 99.1 & 98.2 & 97.4 & 100.0 & 87.6 & 92.6 \\
Std. & 0.11 & 0.17 & 0.24 & 0.00 & 0.21 & 0.29 \\
\bottomrule
\end{tabular}
\normalsize
\caption{Per-run results on the NAVSIM v1.1 \texttt{navtest} split.
NC, DAC, TTC, Comf., EP, and PDMS denote no at-fault collision,
drivable-area compliance, time-to-collision, comfort, ego progress,
and PDM Score, respectively. All values are percentages.}
\label{tab:navsim_seed_results}
\end{table}

\begin{figure*}[!t]
\centering
\includegraphics[width=0.98\textwidth]{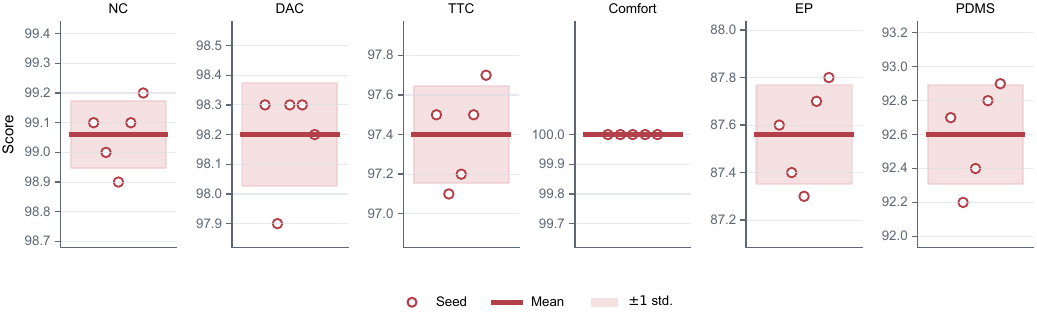}
\caption{Seed-level stability across the six NAVSIM metrics. Open
circles denote independently trained models, horizontal lines denote
arithmetic means, and shaded bands denote one sample standard
deviation across five runs. Exact values are reported in
Table~\ref{tab:navsim_seed_results}.}
\label{fig:navsim_seed_stability}
\end{figure*}

Figure~\ref{fig:navsim_seed_stability} visualizes the same run-level
variation. Open circles denote independently trained models,
horizontal lines denote arithmetic means, and shaded bands denote one
sample standard deviation. PDMS ranges from 92.2 to 92.9, with a
sample standard deviation of 0.29. The improvement reported in the
main paper is therefore not attributable to a single favorable run.

\subsection{Temporal Qualitative Analysis}
\label{sec:supp_navsim_temporal}

We provide consecutive planning frames from three complete NAVSIM
scenes. For each scene, we
select a five-frame excerpt in which the trajectory geometry is
legible. NAVSIM observations are sampled at 2~Hz, so adjacent displayed
frames are separated by 0.5~s and each excerpt spans 2.0~s from its
first to last frame. The yellow rectangle denotes the ego vehicle,
dark-blue rectangles denote surrounding agents, orange denotes the
logged expert trajectory, and cyan denotes the trajectory selected by
\ours.

Each prediction is generated from the observation at its corresponding
logged state. Consequently, these excerpts evaluate whether replanning
remains geometrically consistent across nearby observations. They are
not closed-loop rollouts of the cyan trajectories, and they do not
establish a causal effect of one published trajectory on the next
logged state.

\begin{figure*}[!t]
\centering
\includegraphics[width=0.94\textwidth]{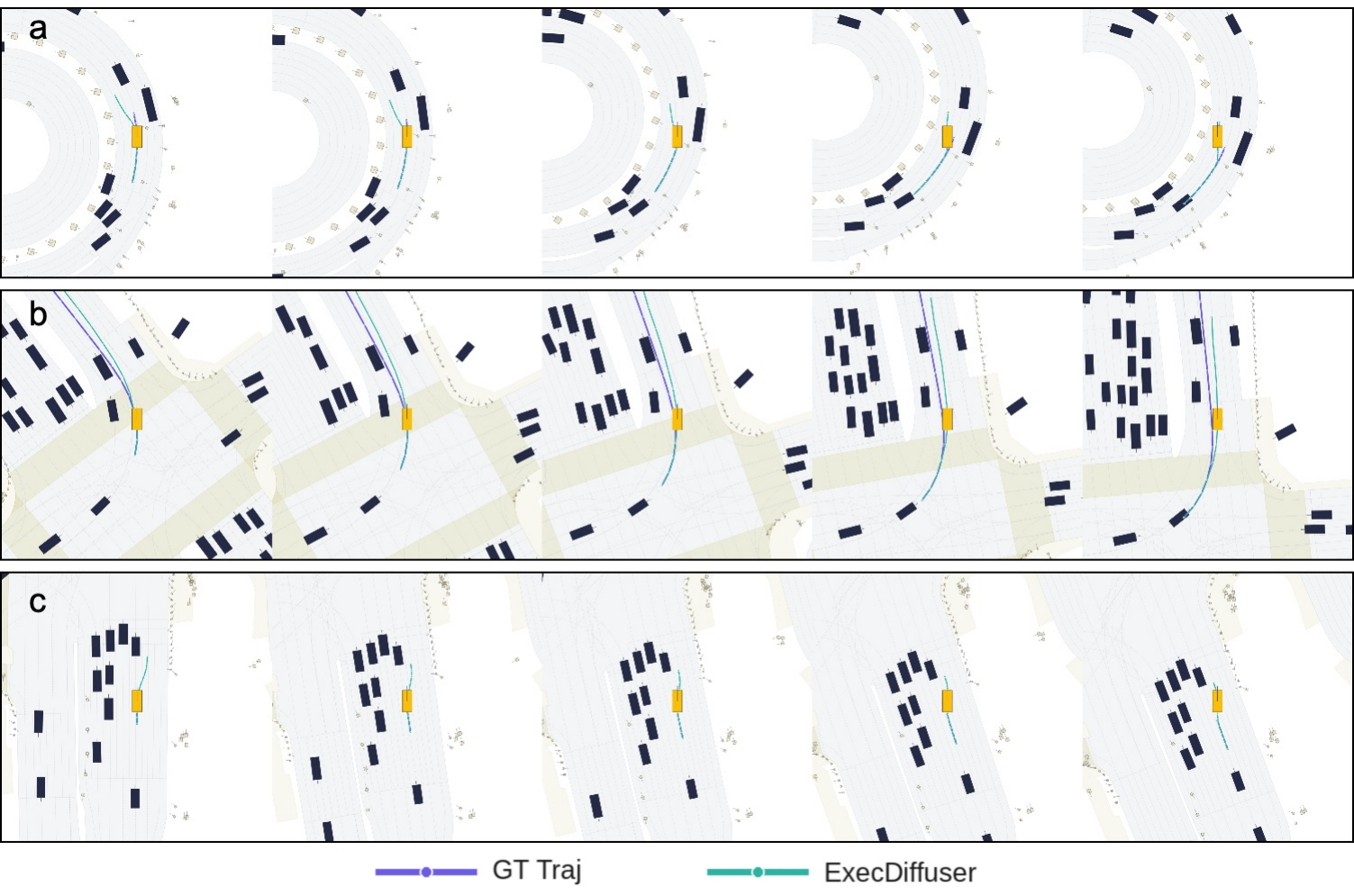}
\caption{Five-frame BEV excerpts from three NAVSIM scenes. Time
progresses from left to right at 0.5-s intervals,
covering 2.0~s from the first to the last displayed frame.
(a) Open-space selection in a left-lane-change context.
(b) Lane-aligned motion after a left turn.
(c) A rightward maneuver around queued traffic.
Orange and cyan denote the logged expert and \ours trajectories,
respectively. The sequences illustrate replanning consistency across
consecutive logged states rather than closed-loop execution of the
predicted trajectories.}
\label{fig:supp_navsim_temporal}
\end{figure*}

In Fig.~\ref{fig:supp_navsim_temporal}(a), \ours repeatedly uses the
more open region on the left while maintaining clearance from traffic
and roadside obstacles on the right. As the logged scene progresses,
the cyan trajectory preserves the same maneuver-level intent instead
of oscillating between the open region and the constrained side. A
short logged-expert trajectory in an individual frame can represent a
stopping decision and is not necessarily missing data.

Figure~\ref{fig:supp_navsim_temporal}(b) shows the intersection exit
after a left turn. Across the five displayed observations, the
selected trajectories remain aligned with the current lane and retain
lateral separation from nearby traffic. Their endpoints and local
curvature change with the observed state, while the underlying
lane-following intent remains stable.

In Fig.~\ref{fig:supp_navsim_temporal}(c), the selected trajectories
maintain a rightward tendency toward less congested space beside the
vehicle queue. The persistence of this direction across the displayed
observations indicates coherent replanning rather than an isolated
single-frame deviation. Together, the three excerpts provide
qualitative evidence of local replanning consistency under evolving
logged observations. Quantitative claims remain grounded in the
official \texttt{navtest} evaluation and the per-run results in
Table~\ref{tab:navsim_seed_results}.


\end{document}